\documentclass[reprint,NumberedRefs]{JASAnew} 
\usepackage{graphicx}
\usepackage{algcompatible}
\usepackage{algpseudocode}
\usepackage{appendix}
\usepackage{upgreek}
\usepackage{amsmath}
\usepackage{array}
\usepackage{amsfonts}
\DeclareMathOperator*{\argmin}{arg\,min}
\DeclareMathOperator*{\argmax}{arg\,max}
\DeclareMathOperator*{\median}{median\,}
\usepackage{bbm}
\usepackage{multirow}

\newcommand*{\citen}[1]{%
  \begingroup
    \romannumeral-`\x 
    \setcitestyle{numbers}%
    \cite{#1}%
  \endgroup   
}

\graphicspath{{Figures/}}

\begin{document}

\title[JASA]{Deep embedded clustering of coral reef bioacoustics}
\author{Emma Ozanich, Aaron Thode, Peter Gerstoft}
\affiliation{Scripps Institution of Oceanography, UCSD}
\author{Lauren A Freeman, Simon Freeman}
\affiliation{Naval Undersea Warfare Center Newport}
\date{\today}

\begin{abstract}
Deep clustering was applied to unlabeled, automatically detected signals in a coral reef soundscape to distinguish fish pulse calls from segments of whale song. Deep embedded clustering (DEC) learned latent features and formed classification clusters using fixed-length power spectrograms of the signals. Handpicked spectral and temporal features were also extracted and clustered with Gaussian mixture models (GMM) and conventional clustering. DEC, GMM, and conventional clustering were tested on simulated datasets of fish pulse calls (fish) and whale song units (whale) with randomized bandwidth, duration, and SNR. Both GMM and DEC achieved high accuracy and identified clusters with fish, whale, and overlapping fish and whale signals. Conventional clustering methods had low accuracy in scenarios with unequal-sized clusters or overlapping signals. Fish and whale signals recorded near Hawaii in February--March 2020 were clustered with DEC, GMM, and conventional clustering. DEC features demonstrated the highest accuracy of 77.5\% on a small, manually labeled dataset for classifying signals into fish and whale clusters.
\end{abstract}
 
\maketitle
\section{Introduction}
 Deep learning, a powerful, recent subfield of machine learning,\cite{deeplearning} excels at learning representations of large amounts of data and often outperforms shallower machine learning methods.\cite{BiancoReview, Niu2019, Huang2018, wang2019transfer, Ozanich2020, Frederick2020, Cao2021} Deep convolutional networks have been particularly effective for image classification due to their scalable capacity. In ocean bioacoustics, machine learning has become an effective signal detection and classification tool,\cite{smirnov2013, Mellinger97, Steiner81, McCowan95, deecke2006, Roch2011, halkias2013, Frasier2016, Frasier2017, Malfante2017, stowell2014, lee2020compact, Bermant2019, Shiu2020, zhong2020, kirsebom2020, berger2019, BiancoReview} with research encompassing both supervised methods,\cite{smirnov2013, Mellinger97, Steiner81, McCowan95, deecke2006, Malfante2017} in which reliable labeled data are present, and unsupervised methods,\cite{Frasier2016, Frasier2017,stowell2014, Lin2018} when labeled data are limited or unavailable. Deep convolutional learning of marine mammal signals in the time-frequency domain has shown promising results for supervised detection and classification.\cite{lee2020compact, Bermant2019, Shiu2020, zhong2020, kirsebom2020, berger2019, halkias2013} For other ocean bioacoustic signals, particularly marine fishes, just a few studies have considered deep learning classification.\cite{Lin2018, Ibrahim2018, Ibrahim2018_featureclassifiers} This study proposes a variant of deep convolutional learning dubbed deep clustering as a method for time-frequency representation learning and classification of unlabeled, automatically detected signals in a cacophonous coral reef environment, characterized by significant presence of spatially dense calling fish.

Coral reefs are amongst the most biodiverse ecosystems in the ocean but are under threat from global climate change, overfishing, and pollution.\cite{Roberts, knowlton, Hoegh-Guldberg, Hughes} Passive acoustic studies allow non-invasive study of reef ecology over continuous time scales of days to years, and offer a complement to measurements that traditionally have been collected through direct observation by SCUBA divers or snorkelers, taking point measurements in time.
\cite{Rogers} Coral reef ambient biological sound, or soundscapes, are an emerging topic of interest in the coral reef scientific community. Reef sound has been linked to relative ecosystem health, abundance of both reef building coral and fleshy macroalgae, and fish density.\cite{Freeman2016, Elise} Healthy reef sounds have also been shown to enhance larval recruitment,\cite{Gordon} suggesting that reef soundscapes are not just a byproduct of biological activity but an integral part of complex ecosystem function. The biological components of reef soundscapes comprise of inadvertent sounds from organism activities such as feeding, locomotion,\cite{Freeman2014} and photosynthesis,\cite{Freeman2018} and acoustic communication such as the fish calls discussed here. Acoustic classification of individual fish calls informs our understanding of spatial and temporal movement, species assemblage, and response to human activities.

Acoustic classification of tropical marine fishes, such as damselfish (family {\it Pomacentridae}), has been improved through passive acoustic field experiments\cite{MannLobel1997, Maruksa2007, TricasBoyle2014reeffish} but lacks established terminology across studies
.\cite{Amorim2006} To reduce the labeling burden, a few studies have considered automatic classification of fish calls by utilizing machine learning tools. \cite{Malfante2017, Lin2018} Unsupervised methods such as Gaussian mixture models (GMM) showed improved detection of fish chorusing events compared to the conventional spectral energy detector.\cite{Lin2018} Deep neural networks, including convolutional networks (CNN), recurrent networks (RNN),\cite{Ibrahim2018} and sparse autoencoders (SAE)\cite{Ibrahim2019_SAE} have achieved higher classification accuracy of grouper calls than a sparse feature classification method.\cite{Ibrahim2018_featureclassifiers} 

In this study, deep clustering is applied to automatically detected, unlabeled signals, combining the benefits of deep learning with the flexibility of clustering. 
%
  Due to the complexity of the dataset, clustering is restricted to two distinct signals: low-frequency fish pulse trains (fish) and segments of whale song units (whale). Standard gaussian mixture model-based clustering and other conventional clustering of handpicked features are conducted for comparison. 
\begin{enumerate}
    \item Spectral and time domain features were manually chosen, or handpicked, based on their observed relation to coral reef fish calling and on studies of fish calling spectral and temporal properties.\cite{MannLobel1997, TricasBoyle2014reeffish, Maruksa2007} Clustering of events using handpicked features provides physical intuition about the event signals but is difficult due to the varying feature properties. Thus, the feature extraction and classification steps are combined into one algorithm using deep learning.
    \item Fixed-length spectrograms were used in deep embedded clustering (DEC),\cite{XieDEC} a deep--learning image compression algorithm that produces an accurate image reconstruction. 
    The DEC latent features were directly clustered with GMM, then the DEC was trained further using a joint clustering loss to encourage deep cluster formation among the latent features.\cite{GuoDEC, Snover, jenkins2021}  
\end{enumerate}
%
Simulated signals were used to compare the limitations of the two approaches before applying them to data recorded on a Hawaiian coral reef between February and March 2020.\cite{Thode_beamform} 

In Sec.~\ref{sec:Methods}, the clustering methods are overviewed and metrics are presented for measuring classification success. 
Section~\ref{sec:SimData} outlines the simulated signals used for method comparison. Experimental data collection and automatic signal detection on a Hawaiian coral reef in Februrary--March 2020 are discussed in Sec.~\ref{sec:RealData} along with discussion of handpicked feature extraction. A comparison of clustering results on the simulated datasets are presented in Sec.~\ref{sec:SimResults}, with an analysis of the effect of the DEC latent dimension. Finally, Sec.~\ref{sec:RealResults} presents DEC, GMM, and conventional clustering of two signal types 
for the experimentally 
detected signals.  Section~\ref{sec:discussion} summarizes the approach and discusses challenges associated with the methods and dataset.

\section{Methodology\label{sec:Methods}}
Clustering methods are unsupervised frameworks for categorizing data according to their similarities.\cite{BeimannCh4} This section describes deep embedded clustering for learning the latent feature vectors and forming clusters directly from high-dimensional data. For comparison, the Gaussian mixture models, K-means, and hierarchical agglomerative clustering algorithms, common methods that rely on approximations of the input feature vector properties, are also presented.

\subsection{Deep embedded clustering}
\begin{figure*}[bt]
    \centering
    \includegraphics[width = 0.95\textwidth]{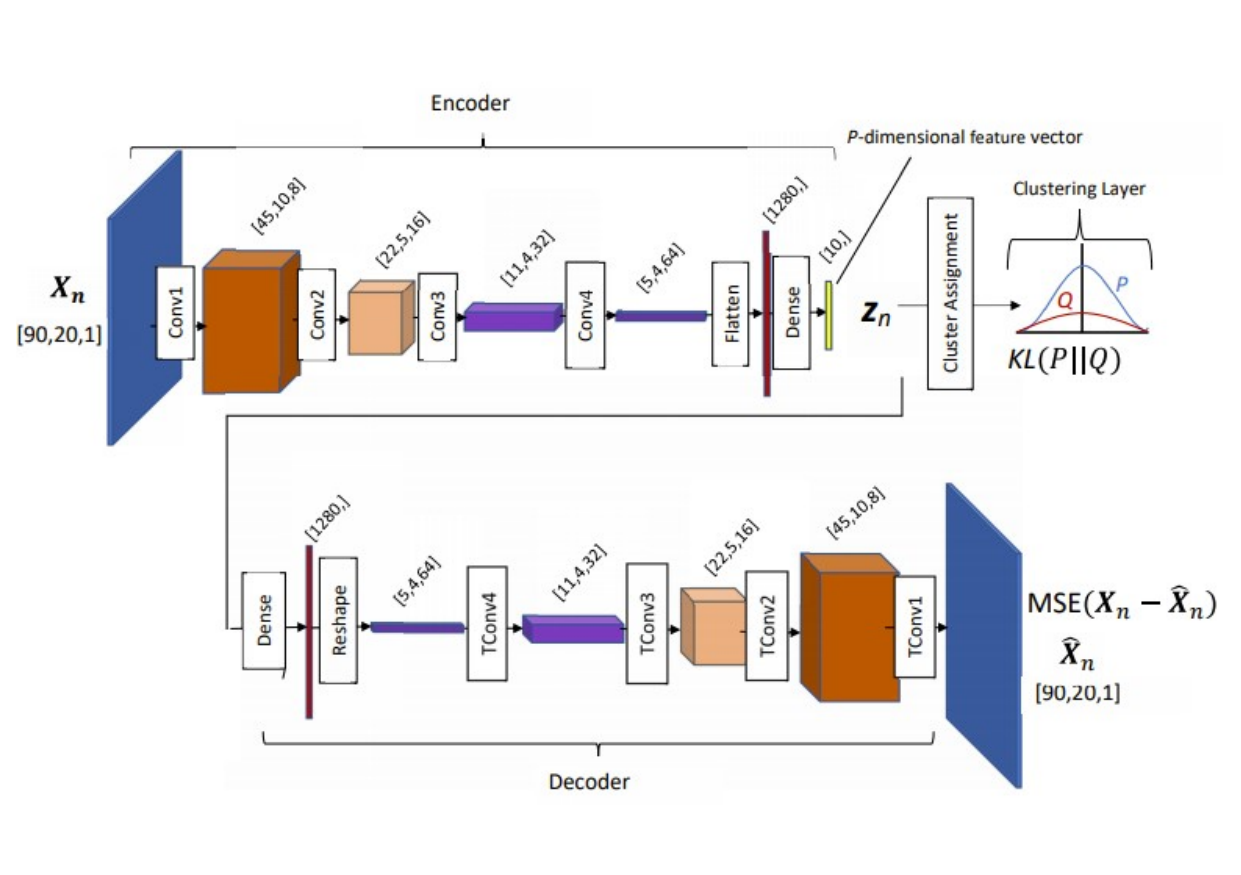}
    \caption{Architecture of the deep embedded encoding (DEC) convolutional model. Convolutional filters were sized 3x3 and used the rectified linear unit (ReLU) activation function.}
    \label{fig:DEC}
\end{figure*}

Deep embedded clustering (DEC) is a modified convolutional autoencoder, a deep feature learning method,\cite{Goodfellow_AE} that uses a joint loss function to transform a set of $N$ input data, $\cal{X}=\{\mathbf{X}_{\rm \it 1},\dots,\mathbf{X}_{\rm N}\}$, $\mathbf{X}_n\in\mathbb{R}^{d_1\times d_2}$, into a set of latent feature vectors, $\cal{Z}=\{\mathbf{z}_{\it \rm 1},\dots,\mathbf{z}_{\rm N}\}$, $\mathbf{z}_n\in\mathbb{R}^P$. 
The output is a reconstruction of the input data, $\widehat{\bm X}_n\in\mathbb{R}^{d_1 \times d_2}$, estimated using the latent feature vectors. 
The DEC structure consists of two stacked networks (Fig.~\ref{fig:DEC}): the encoder network, which maps input data into a lower-dimensional latent space, and the decoder network, which reconstructs an approximation of the input from the latent space. For a given latent feature dimension, $P$, DEC learns representative latent features using the mean squared reconstruction error of the inputs,
 \begin{align}
    MSE &= \frac{1}{N}\sum_{n=1}^N |\mathbf{X}_n-\widehat{\mathbf{X}}_n|^2.
\end{align}
Inspired by recent applications to spectrograms of unlabeled seismic events,\cite{Snover, jenkins2021} the compressive architecture 
downsamples the input using convolutions of stride length 2. The Rectified Linear Unit (ReLU) was used to transform the outputs at each layer. 
The neural network structure\cite{Snover} 
was adapted by reducing the convolutional filter size 
and adding convolutional layers to downsample the input. 

The DEC weights were pretrained for 1000 epochs using the MSE of fixed-length spectrogram images containing $N_f$ frequencies and $N_t$ time samples, $\mathbf{X}_n$
$\in \mathbb{R}^{N_f\times N_t}$, and their reconstructions, $\widehat{\mathbf{X}}_n\in\mathbb{R}^{N_f\times N_t}$. The Adam optimizer\cite{Kingma} was used with learning rate of $10^{-3}$. The spectrogram length $N_t$ corresponds to a signal duration of $0.5$ s $\!=\!N_t\!\cdot\! dT$ and was motivated by the signals of interest in the coral reef soundscape.  Events longer than $0.5$ s were clipped to length to ensure a fixed size. 

 Each latent feature vector $\bm{z}_n$ is assumed to belong to one of $K$ clusters, $C_k$, with means $\bm{\mu}_k\in\mathbb{R}^P$, $k=1\dots,K$. The latent features $\mathbf{z}_n$ were clustered with GMM to initialize the deep clustering, with means $\pmb{\mu}_k$, $k\!=\!1,\dots,K$. These clustering results are reported as ``{\it GMM, latent features}" for the simulated and experimental data.
 
Last, the DEC latent feature vectors were updated by incorporating the Kullback-Leibler (KL) divergence into the training loss. DEC was trained for an additional 20 epochs using the joint clustering/reconstruction loss function,
\cite{GuoDEC, XieDEC}
\begin{align}
    Loss &= 0.1\cdot KL + 0.9\cdot MSE,\label{eq:KL}\\     KL &=\sum_n^N \sum_k^K p_{nk}\log\left(\frac{p_{nk}}{q_{nk}}\right),\\
    q_{nk} &= \frac{(1+\|\mathbf{z}_n-\pmb{\mu}_k\|^2)^{-1}}{\sum_j (1+\|\mathbf{z}_n-\pmb{\mu}_j^2\|^{-1}}\label{eq:qnk},\\
    p_{nk} &= \frac{q_{nk}^2/\sum_m q_{mk}}{\sum_j (q_{nj}^2/\sum_m q_{mk})}\label{eq:pnk},
\end{align}
 \eqref{eq:qnk} is the empirically estimated Student's t-distribution of the latent feature vectors around each cluster mean, and \eqref{eq:pnk} further penalizes points that are distant from cluster centers.\cite{GuoDEC} The objective of additional training is to learn latent feature vectors that are both disjoint in feature space and representative of the inputs. The weights for each loss term in \eqref{eq:KL} are from Ref.~\citen{GuoDEC}.
Details of the model are given in Fig.~\ref{fig:DEC} and Table~\ref{tab:DEC_structure}.  
The DEC was written in Keras\cite{chollet2015keras} using Tensorflow\cite{tensorflow2015-whitepaper}. 

\begin{table*}[tb]
    \centering
    \scriptsize
    \caption{\label{tab:DEC} Network architecture used in deep embedded clustering. The input and output shapes are given as [height, width, depth]. The kernel size is the shape of the two dimensional convolutional filters in [height, width].}
    \begin{tabular}{p{1.2cm}p{2.7cm} p{1.5cm} p{1cm} p{1cm} p{1cm} p{1.6cm} p{1.5cm} p{1.8cm}}
        \hline\hline
       Layer Name & Layer Type & Input shape  & Filters & Kernel size & Stride & Activation & Output shape & Parameters  \\ \hline 
         Conv1 &  2D Convolution & [90,20,1] & 8 & [3,3] & [2,2] & ReLU & [45,10,8] & 80\\
       &  2D Conv. & [45,10,8] & 8 & [2,1] & [1,1] & ReLU & [44,10,8] & 136\\
       Conv2 &  2D Conv. & [44,10,8] & 16 & [3,3] & [2,2] & ReLU & [22,5,16] & 1168\\
    & 2D Conv. & [22,5,16] & 16 & [1,2] & [1,1] & ReLU & [22,4,16] & 528\\
    Conv3 & 2D Conv. & [22,4,16] & 32 & [2,1] & [2,1] & ReLU & [11,4,32] & 1056\\
     & 2D Conv.  & [11,4,32] & 64 & [2,1]  & [1,1] & ReLU & [10,4,64] & 4160\\
    Conv4 &  2D Conv. & [10,4,64] & 64 & [2,1] & [2,1] & ReLU & [5,4,64] & 8256\\
     Flatten &  Flatten & [5,4,64] & - & - & - & - & [1280] & 0\\
     Encoded & Fully Connected & [1280] & - & - & - & ReLU & [10] & 6405 \\
     Dense &  Fully Connected & [15] & - & - & - & ReLU & [1280] & 7680 \\
     Reshape &  & [1280] & - & - & - & - & [5,4,64] & 0\\
     TConv4 & Transposed &&&&&&&\\
     & Convolution & [5,4,64] & 32 & [2,1] & [2,1] & ReLu & [10,4,32] & 4128  \\
      & T. Conv. & [10,4,32] & 32 & [2,1] & [1,1] & ReLu & [11,4,32] & 2080 \\
      TConv3 & T. Conv. & [11,4,32] & 16 & [2,1] & [2,1] & ReLu & [22,4,16] & 1040\\
      & T. Conv. & [22,4,16] & 16 & [1,2] & [1,1] & ReLu & [22,5,16] & 528\\
    TConv2 & T. Conv. & [22,5,16] & 8 & [3,3] & [2,2] & ReLu & [44,10,8] & 1160\\
      & T. Conv. & [44,10,8] & 8 & [2,1] & [1,1] & ReLu & [45,10,8] & 136\\
    TConv1 & T.  Conv. & [45,10,8] & 1 & [3,3] & [2,2] & Linear & [90,20,1] & 73\\
             \hline\hline
    \end{tabular}

    \label{tab:DEC_structure}
\end{table*}

\subsection{Gaussian Mixture Models}
Gaussian mixture models (GMM) aim to partition a set of $N$ feature vectors, $\cal{X}=\{\mathbf{x}_{\it \rm 1},\dots,\mathbf{x}_{\it N}\}$ into $K$ clusters, $C_k$, $k=1,\dots,K$, assuming the $k$th cluster is Gaussian with prior probability $\pi_k$, mean $\bm{\mu}_k$, and covariance $\bm{\Sigma}_k$.\cite{Murphy_cluster} $C_k$ is a nonempty set. If the number of clusters, $K$, is assumed {\it a priori}, then the Gaussian parameters $\bm{\theta}_k = (\pi_k, \bm{\mu}_k, \bm{\Sigma}_k)$ can be iteratively estimated.

GMM employs the Expectation-Maximization (EM) algorithm to estimate $\bm{\theta}_k$ from $\cal{X}$ via the complete data likelihood.\cite{Murphy_cluster}
EM is an alternating algorithm that updates the weighted posterior probability (responsibility, $r_{nk}$) and the prior, mean, and covariance by\cite{Murphy_cluster}
\begin{align}
   &\text{E step:} \quad r_{nk}\!=\! \frac{\pi_k p(\mathbf{x}_n|\bm{\theta}_k^{t-1})}{\sum_{k'=1}^K \pi_{k'}p(\mathbf{x}_n|\bm{\theta}_k^{t-1})},\label{eq:Estep}\\
    &\text{M step:}  \quad\pi_k \!=\! \frac{1}{N}\sum_{n=1}^N r_{nk}, \quad k=1,\dots,K \\
&\bm{\mu}_k \!=\! \frac{\sum_{n=1}^N r_{nk}\mathbf{x}_n}{\sum_{n=1}^N r_{nk}}, \quad
\bm{\Sigma}_k = \frac{\sum_{n=1}^N r_{nk}\mathbf{x}_n \mathbf{x}_n^T}{\sum_{n=1}^N r_{nk}} - \bm{\mu}_k\bm{\mu}_k^T\!\!
\end{align}
for $\mathbf{x}_n \in \cal{X}$, $n=1,\dots,N$. $\bm{\theta}_k = \{\pi_k, \bm{\mu}_k, \bm{\Sigma}_k\}$ represents the estimated parameters of the $k$th cluster at the $t$th iteration. 
In GMM, the data likelihood $p$ is Gaussian, 
\begin{align}
    p(\mathbf{x}_n|\pmb{\theta}_k^t) &\propto e^{-\frac{1}{2}(\mathbf{x}_n-\bm{\mu}_k)^T\bm{\Sigma}_k^{-1}(\mathbf{x}_n-\bm{\mu}_k)}
\end{align}
where $\propto$ represents proportionality. EM is iterated until $\pi_k$, $\bm{\mu}_k$, and $\bm{\Sigma}_k$ converge for all $k$. 
The estimated covariance matrix can become ill-conditioned if there are fewer than $K$ true clusters.

The cluster priors $\pi_k$, means $\bm{\mu}_k$, and covariance $\bm{\Sigma}_k$, $k=1,\dots,K$ were initialized for 100 starts using the K-means++ algorithm (App.~\ref{appendix}).\cite{k++}

\subsection{K-Means}
 K-means\cite{ESL} is an approximation to the EM algorithm 
 that partitions a set of data-derived feature vectors $\cal{X}=\{\mathbf{x}_{\rm \it 1},\dots,\mathbf{x}_{\rm \it N}\}$ into $K$ clusters, $C_k$, $k=1,\dots,K$, where $C_k$ is a nonempty set. The number of clusters $K$ must be assumed {\it a priori}. The K-means algorithm is derived assuming the feature vectors belonging to the $k$th cluster are drawn from a Gaussian distribution with covariance $\pmb{\Sigma} = \sigma^2 \mathbb{I}$ and prior probability $\pi_k = 1/K$, where $\mathbb{I}$ is the identity matrix,\cite{Bishop}


\begin{align}
r_{nk} &= \begin{cases}1 & \text{if } k=\argmin_{j} \|\mathbf{x}_n \!-\! \pmb{\mu}_j\|_2^2,\!\! \quad\!\!\! j=1,\dots,K\!\!\\ 0 & {\rm otherwise}, \end{cases}\nonumber\\
\pmb{\mu}_k &=  \frac{1}{|C_k|}\sum_{\mathbf{x}_n \in C_k} \mathbf{x}_n = \frac{\sum_n r_{nk}\mathbf{x}_n}{\sum_n r_{nk}}\label{eq:kmeans_update}
\end{align}
for $n=1,\dots,N$. $|C_k|$ denotes the cardinality or size of the $k$th cluster. 

The cluster means $\pmb{\mu}_j$, $j=1,\dots,K$ were initialized for 100 starts using the K-means++ algorithm (App.~\ref{appendix}).\cite{k++}

K-means is commonly used due to its computational speed and interpretability, but its validity is subject to the isotropic covariance assumption and the equality of the cluster priors. 
If the true number of classes differs from $K$, the clusters will be incorrectly estimated. 
\subsection{Agglomerative hierarchical clustering}

Agglomerative hierarchical clustering,\cite{BeimannCh4, Murphy_cluster} also called bottom-up clustering, partitions a set of $N$ data-derived feature vectors, $\cal{X}=\{\mathbf{x}_{\rm \it 1},\dots,\mathbf{x}_{\rm \it N}\}$, into $K$ clusters $C_k$, $k=1,\dots,K$, by grouping the most similar data at each step. Hierarchical clustering  successively merges nearby clusters until the stop criterion is achieved. In this case, the stop criterion is met when $K$ or fewer clusters remain, where $K$ must be set by the practitioner.

Agglomerative clustering does not require the number of clusters to be assumed in advance. 
As demonstrated with classification of dolphin echolocation clicks, \cite{Frasier2016, Frasier2017} bottom-up methods allows for detailed clusters to be sequentially merged as additional information or labels become available.

In agglomerative clustering, the $k$th cluster at the $m$th iteration is represented as $C_{k_m}$, $k_m=1,\dots,N\!-\!m\!+\!1$. The number of clusters is decreased to $N\!-\!m$ by merging the two closest clusters that satisfy 
\begin{align}
   j,k \!&=\! \argmin_{i,{i'}} d(C_i,C_{i'}), \!\!\quad i, i'= 1,\dots,N\!-\!m\!+\!1, i\neq i' \nonumber\\
  C_{k'} \!&=\! \{C_j \cup C_k\},\quad \!\!\!
  \!C_{k_{m+1}} \!\!=\! \{C_{k_m},\! C_{k'}\} \!-\! \{C_j\}\!-\!\{C_k\!\}  \label{eq:merge}
\end{align}
 $d(C_i,C_{i'})$ is a metric measuring the distance between all points in sets $i$ and $i'$. 
 This agglomerative process is repeated $N\!-\!K+1$ times, until there are at most $K$ clusters $C_{k}$ remaining, with $k = 1,\dots,K$.

The distance metric $d$ chosen here is from  {\it Ward's method}.\cite{Ward}
Ward's method measures the incremental sum-of-squares resulting from merging two sub-clusters.\cite{HierarchicalClustering}   For clusters $C_j$ and $C_{k}$ with cluster means $\pmb{\mu}_j$ and $\pmb{\mu}_k$ that were merged for form $C_{k'}$ with mean $\pmb{\mu}_{k'}$, the Ward's method distance is
\begin{align}\label{eq:sse}
    d(C_j,C_k) &= \!\!\!\!\sum_{\mathbf{x}_i\in C_{k'}}\!\!\!(\mathbf{x}_i\!-\!\pmb{\mu}_{k'})^2\!-\!\!\!\!\!\sum_{\mathbf{x}_{i'}\in C_j}\!\!(\mathbf{x}_{i'}\!-\!\pmb{\mu}_j)^2\!-\!\!\!\!\!\sum_{\mathbf{x}_{i''}\in C_k}\!\! (\mathbf{x}_{i''}\!-\!\pmb{\mu}_k)^2 \nonumber \\
    &= \frac{2|C_j| |C_{k}|}{|C_j|+|C_{k}|}\|\pmb{\mu}_j - \pmb{\mu}_{k}\|_2^2,
\end{align}
$|C_j|$ and $|C_k|$ are the cardinality of cluster $C_j$ and $C_k$. 
Ward's method has been empirically successful,\cite{EverittCh4} but has been shown to perform worse in cases with unequal-sized clusters.\cite{HandsEveritt1987} Single-linkage, centroid, and complete-linkage methods were also considered, but had low accuracy. 


\subsection{Metrics\label{subsec:metrics}}

 The performance of the handpicked feature clustering and DEC was measured by 
\begin{align}\label{eq:accuracy}
    \text{Accuracy}&= \frac{1}{N}\sum_{i=1}^N \mathbb{I}(t_i,\hat{t}_i), \quad t_i, \hat{t}_i \in \{0,1,2\} \\
    \text{Precision} &\!=\! \frac{\sum_{i=1}^N \mathbb{I}(t_i, \hat{t}_i)\mathbb{I}(t_i, 0)}
    {\sum_{j=1}^N \mathbb{I}(\hat{t}_j, 0)},\\ 
    \text{Recall} \!&=\! \frac{\sum_{i=1}^N \mathbb{I}(t_i, \hat{t}_i)\mathbb{I}(t_i, 0)}
    {\sum_{j=1}^N \mathbb{I}(t_j, 0)},
\end{align}
where $t_i$ is the true label and $\hat{t}_i$ is the categorical prediction. The indicator function $\mathbb{I}(x) = 1$ if $x$ is True and $\mathbb{I}(x) = 0$ otherwise. Here, whale (label 0) was the true category. Fish were indicated by label 1. The label 2 was used in a scenario to indicate overlapping fish/whale signals. Precision (positive predictive value or PPV) measured the ratio of correctly predicted whale to total predicted whale. Recall (hit rate or true positive rate) measured the ratio of correctly classified whale to the true total. Higher metrics correspond to improved performance, with perfect performance given by accuracy, precision, and recall all equal to 1.


\section{Simulated datasets\label{sec:SimData}}

A set of simulated coral reef bioacoustic signals, motivated by observed signals from a coral reef soundscape near Hawaii, was simulated to compare conventional clustering of handpicked features to deep embedded clustering. 

\begin{figure}[tb]
\centering
\includegraphics[width=0.5\textwidth]{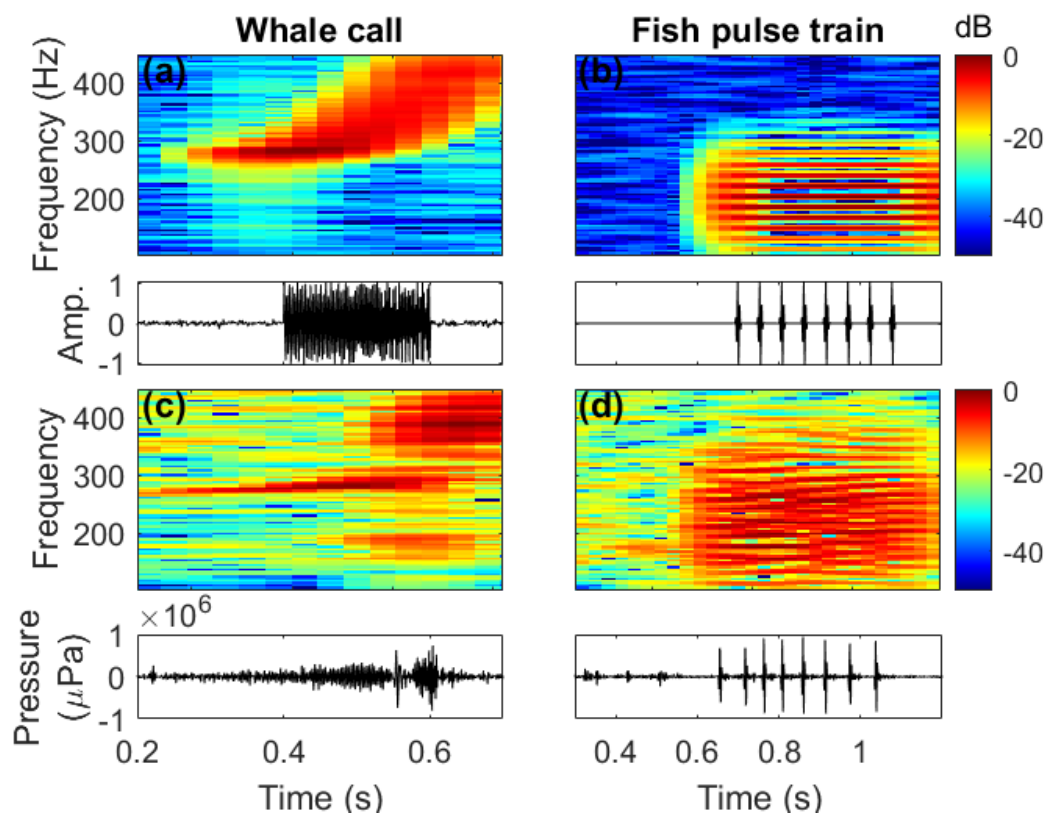}
\caption[Simulated (a--b) and measured (c--d) coral reef bioacoustic signals before clipping.]{\label{fig:calls}  Simulated (a--b) and measured (c--d) coral reef bioacoustic signals before clipping. Whales were simulated with an FM upsweep (a,c), and fish were simulated with superimposed Gaussian pulses (b,d). White noise was added to both signals at 25 dB SNR relative to the mean signal power.}
\end{figure}

One hundred simulation trials were conducted, with 
10,000 signals simulated for each trial. The signals were simulated as timeseries sampled at 1 kHz and mimicked whale and fish signals (Fig.~\ref{fig:calls}). A spectrogram image of each signal was generated using a 256--pt FFT with 90\% overlap. The fixed-length spectrograms were created by clipping all signals to 0.5 s. 

The DEC latent feature dimension was optimized over a range of values (Fig.~\ref{fig:hyperparameter}). The optimal DEC performance was compared to clustering with automatically extracted handpicked features. Then, a third category containing an overlapping fish and whale signals was included, with 33\% of the total samples belonging to each category.

Quadratic FM sweeps mimic parts of humpack whale song units (whale). The equation for the instantaneous frequency of a quadratic sweep is\cite{gausspuls}
\begin{align}
    f(t) &= \beta t^2 + 2\pi f_0, \quad \beta = \frac{2\pi\Delta f}{T}
\end{align}
where $t$ is the continuous time variable, $f_0$ is the initial sweep frequency, $\Delta f$ is the total bandwidth, and $T$ is the duration of the signal. The signal is an FM upsweep when $\beta>0$, an FM downsweep when $\beta<0$ and a tonal when $\beta=0$. The phase $\Phi(t)$ of the time domain signal is found by integrating the instantaneous frequency\cite{gausspuls}
\begin{align}
    \Phi(t) &= \int_0^{t-t_0} (\beta \tau^2 + 2\pi f_0) d\tau \\
    y(t) &= \sin(\Phi(t)) = \sin(\frac{\beta}{3}(t-t_0)^3 + 2\pi f_0(t-t_0)),
\end{align}
where $t_0$ is the signal start time. 
For simulation, $t$ is discretized at 
$0.001$ s 
(sampling frequency of 1 kHz).

Timeseries of impulses, or pulse trains, simulated fish pulse calling (fish). The pulses were a set of $N_p$ superimposed Gaussian-modulated sinusoids\cite{chirplet} spaced $\Delta t$ apart
\begin{align}
    y(t) &= \sum_{i=0}^{N_p-1} e^{-a|t-i\Delta t-t_0|^2}\sin(2\pi f_c(t-i\Delta t-t_0))\\
    a \!&=\!  \tau^{-2}2\log(2), \nonumber
\end{align}
where $\tau$ is the half-power pulse width and $f_c$ is the center frequency.

The signal parameters were varied randomly for each sample (Table~\ref{tab:params}). Pulse width and center frequency were fixed to achieve a representative pulse structure. The number of pulses and spacing were drawn from experimentally estimated distributions. The duration, initial frequency, and total bandwidth of the FM sweep were drawn at uniform random from a range of realistically observed values. All signals were centered within the 0.5 s spectrogram and assigned a random delay of within $\pm0.1$ s.

\begin{table}[tb]
\caption{\label{tab:params} Signal parameters were drawn from random distributions for each simulated event.}
\begin{center}
\begin{tabular}{c  c  c}
\hline\hline
& FM Sweep & Pulse train \\
\hline
Duration/Width (s) & $T\in \mathcal{U}(0.2,0.4)$ & $\tau=0.005$ \\
Delay (s) & $t_0\in\mathcal{U}(-0.1,0.1)$ & $t_0\in\mathcal{U}(-0.1,0.1)$ \\
Frequency (Hz) & $f_0 \in\mathcal{U}(100,400)$ & $f_c=200$\\
& $\Delta f\in\mathcal{U}(-150,150)$ & \\
Peak spacing (s) & & $\Delta t \in 0.47$*beta$(4,23)$\\
Number of peaks & & $N_p\in\lfloor 13$*beta$(3.5,8)\rfloor$\\
SNR (dB) & $SNR\in\mathcal{U}(15,30)$ & $SNR\in\mathcal{U}(0,30)$ \\
\hline\hline
\end{tabular}
\end{center}
\end{table}

White noise was added to the simulated signals using a fixed signal-to-noise ratio (SNR),  
\begin{align}\label{eq:snr}
    SNR &= 10\log_{10}\frac{\sigma_s^2}{\sigma_n^2},\\
     y(t) &= y(t)+ \mathcal{N}(0, \sigma_n^2\mathbb{I}),
\end{align}
where $\sigma_s^2$ is the signal power and $\sigma_n^2$ is the noise power. The SNR range of each signal was determined from the experimental spectrograms during manual labeling. The SNR in dB was estimated as the difference of the peak signal power to the median power of the background.

The signal power was estimated as the bandwidth-normalized mean power over the signal duration,\cite{ManolakisProakis}
\begin{align}
    \sigma_{s, FM}^2 \!\!&= \frac{1}{\Delta f\cdot T}\int_{t_0}^{t_0+T} \!\!\!\!\!\!\!\!|y(t)|^2 dt, \\ \sigma_{s, pulse}^2\!\! &= \frac{1}{\Delta f \cdot 4\tau}\int_{t_0-2\tau}^{t_0+2\tau} \!\!\!\!\!\!|y(t)|^2 dt.
\end{align}
Three handpicked features were extracted: peak frequency \eqref{eq:peaks}, kurtosis \eqref{eq:kurtosis}, and number of timeseries peaks \eqref{eq:npeaks}. Duration, median power, and cross-sensor coherence were excluded due to limitations of the fixed simulation parameters. Then, GMM, K-means, and hierarchical clustering were applied to the feature matrix to discover $K\!=\!2$ clusters (fish or whale) or $K\!=\!3$ (fish, whale, or both).


\section{Hawaiian coral reef dataset}\label{sec:RealData}
\subsection{Automatically detected signals in experimental data}
\begin{figure}[tb]
\centering
\includegraphics[width = 0.5\textwidth]{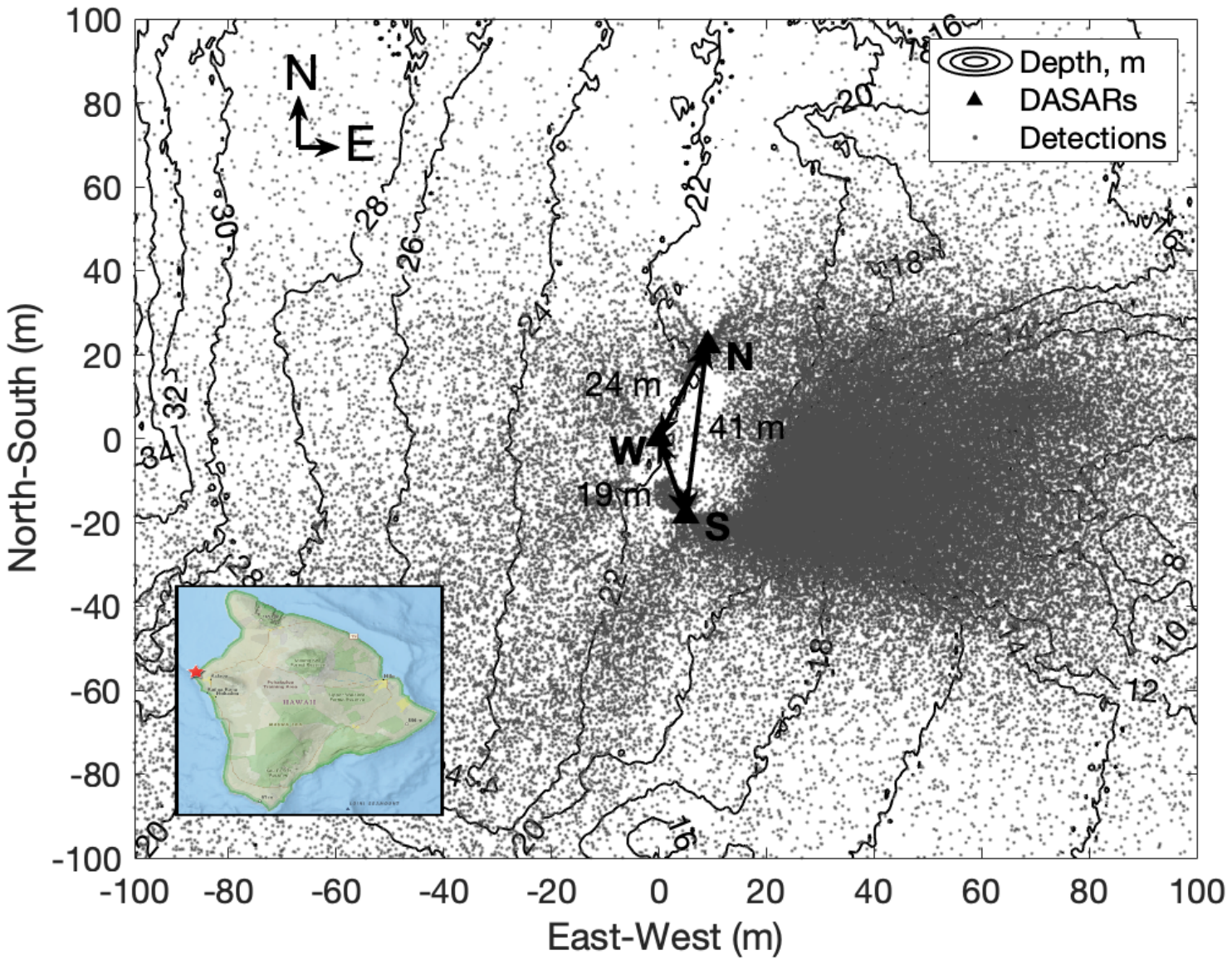}
\caption[Diagram of the DASAR array deployed adjacent to a coral reef on the island of Hawaii.]{\label{fig:array_plot} Diagram of the DASAR array deployed adjacent to a coral reef on the island of Hawaii. The estimated detection locations are shown as gray dots. The majority of the reef is located due east of the array. The sensor positions were measured on the seafloor relative to DASAR W.}
\end{figure}
Three directional autonomous seafloor acoustic recorders (DASARs) were deployed adjacent to a coral reef westward of the island of Hawaii. The DASARs, labeled N, W, and S for the north-most, middle, and south-most sensors,  measured pressure and horizontal particle velocity with $x$-- and $y$-- components oriented at orthogonal compass directions. The array was roughly oriented N-S with inter-sensor spacing about 15 m (Fig.~\ref{fig:array_plot}).

The DASARs recorded continuously for 7 days with a sampling rate of 1 kHz. On February 25, 2020, the dominant soundscape contributors below 500 Hz were reef fish, humpback whales, and motor noise from transiting surface boats, with boat noise occurring predominantly during daylight and fish calling most pronounced during dusk.
The data were processed in 5 minute chunks to account for DASAR clock drift. For each chunk, the pressure spectrograms on the three sensors were cross-correlated and aligned to within a time bin.\cite{Thodeinreview} The spectral density power spectrogram, denoted as matrix $\mathbf{X}\in\mathbb{R}^{N_f\times N_t}$, was generated using a 256--point FFT with 90\% overlap and Hanning window and normalized to have units of $\log \upmu$Pa$\cdot$Hz$^{-1}$, with ${\rm d}t=0.026$ s and ${\rm d}f = 3.91$ Hz. 

A directional event detector\cite{Thodeinreview} was developed to utilize the DASARs' directional capability by combining two DASARs (App.~\ref{app:detector}). 
For this study, the detector parameters were $\Delta \theta = 90^\circ$, $T=120$ Hz, $M_{\rm sep} = 1$ ($M_{\rm sep}\cdot {\rm d}t = 0.0256$ s), and $T_{\rm max}=2$ s. Detected events were localized\cite{Lenth1981} to ensure 
that their signal had sufficient bandwidth for feature extraction. 
 Detections for which the localization algorithm failed to converge were discarded. The remaining events were spatially filtered within a 100 m by 100 m box from DASAR S (Fig.~\ref{fig:array_plot}). 92,736 potentially localizable detections within 100 m were kept for further analysis, on average about 1 detection per second.

\begin{figure}[tb]
\centering
\includegraphics[width = 0.48\textwidth]{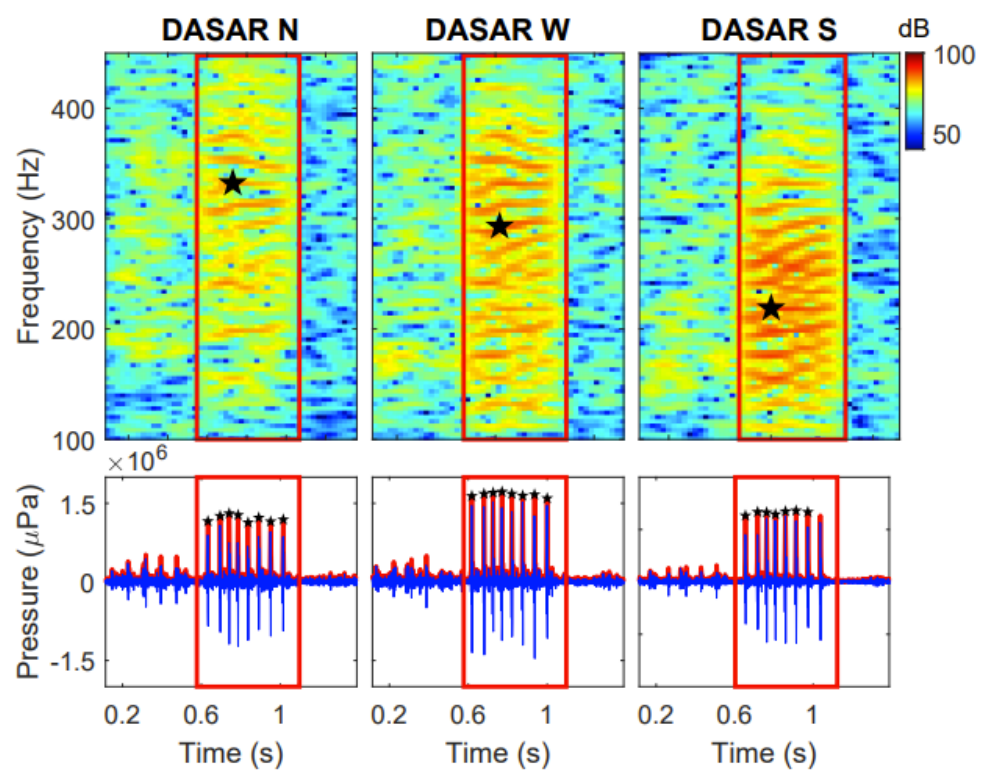}
\caption[Handpicked features of a fish call event on February 25, 07:12 HST, measured on directional autonomous seafloor acoustic recordings (DASARs) N, W, and S.]{\label{fig:feature_extract}Handpicked features of a fish call event on February 25, 07:12 HST, measured on directional autonomous seafloor acoustic recorders (DASARs) N, W, and S at 1 kHz. The spectrograms were generated using a 256--point FFT with 90\% overlap. A call duration of 0.52 s was determined during the detection process. {(a--c)} The spectrograms were used to extract peak frequencies (black star) from 219--332 Hz and median PSD from 72.8--76.3 dB. { (d--f)} The timeseries envelope was used to extract the kurtosis values of 18--21, cross-sensor coherences of 0.72--0.74, and 8 temporal peaks. DASAR S was the closest to the call.}
\end{figure}

\subsection{Spectral feature extraction}

The handpicked features (Table~\ref{tab:features}, Fig.~\ref{fig:feature_extract}) are time-frequency properties known to relate to fish call type including power, duration, and peak frequency\cite{MannLobel1997, Maruksa2007, TricasBoyle2014reeffish} as well as timeseries estimates of impulsive noise.\cite{Martin2020, Malfante2017} 

If $t_{n,1}$ and $t_{n,2}$ represent the absolute start and end of the $n$th detected signal in seconds,
the event duration is
\begin{align}
    {\rm Duration} \hspace{1mm} {\rm (s)} &= \Delta t_n = t_{n,2}-t_{n,1}.
\end{align}
The median power and peak frequency of the $n$th detection were extracted from the spectrogram,
(Fig.~\ref{fig:feature_extract}a--c), with
\begin{align}\label{eq:peaks}
    {\rm Median\ PSD} &= \median_{i,j}X_n^{(i,j)} \nonumber \\
    {\rm Peak\ freq.} &= \argmax_{i} \left( \max_j X_n^{(i,j)}\right), \nonumber \\
    &\quad\quad i\!=\!1,\dots,N_f, \quad j\!=\!1,\dots,N_t. \nonumber
\end{align}
%
%
The spectral features extracted at DASAR S, closest to the majority of detected signals (Fig.~\ref{fig:array_plot}), were used for clustering.

\subsection{Temporal feature extraction}
\renewcommand{\arraystretch}{0.75}
\begin{table}[tb]
    \centering
        \caption{Event features estimated for automatically detected signals.}
\begin{tabular}{l l}
\hline\hline
     Feature Name (units) &  Description \\\hline\\
    Kurtosis & Fourth moment normalized by the\\
    &  squared variance  \\ \\
        Npeaks (count) & Number of peaks with at least 5 \\
        & dB prominence {\it re} the standard \\
        & deviation   \\ \\
    Peak frequency (Hz) & Frequency of the peak power \\
    & spectral density \\ [2ex]
   \multicolumn{2}{c}{{\it Features for experimental data only}}\\\hline\\[0.2ex]
       Duration (s) & Length of detected event\\\\
    Coherence & Normalized time coherence \\
    &between DASARs \\\\
    Median PSD (dB)  & Median power spectral density (PSD)\\
    & across event mask \\ 
    \hline\hline
\end{tabular}
    \label{tab:features}
\end{table}
The pressure timeseries of the $n$th detected signal $\mathbf{y}_n\in \mathbb{R}^{M}$ was extracted between $t_{n,1}$ and $t_{n,2}$ (Fig.~\ref{fig:feature_extract}) and contained $M$ time samples. For the experimental data, the vector sensor x-- and y-- velocity during the $n$th detected signal, $\mathbf{v}_{n,x}\in\mathbb{R}^{M}$ and $\mathbf{v}_{n,y}\in\mathbb{R}^{M}$, were used to create a beamformed pressure timeseries, $\mathbf{y}_b\in \mathbb{R}^M$, for improved detection SNR,
\begin{align}
    \mathbf{y}_b &= \mathbf{y}_n + Z_0\left[\mathbf{v}_{n,x} \sin(\hat\theta) +  \mathbf{v}_{n,y}\cos(\hat\theta)\right], 
\end{align}
where $Z_0=\rho c$ is the impedance in water with density $\rho$ (kg $\!\cdot\! $ m$^3$) and sound speed $c$ (m $\!\cdot\!$ s$^{-1}$).\cite{Thode_beamform}  
$\hat{\theta}$ is the signal azimuth estimated during the directional detector processing.\cite{Thodeinreview} For the simulated data, $\mathbf{v}_{n,x}=\mathbf{v}_{n,y}=\mathbf{0}$. 

Three metrics were chosen for timeseries extraction: kurtosis, number of peaks, and cross-sensor coherence. The kurtosis is a ratio of moments and has recently been applied to the task of differentiating impulsive from non-impulsive sounds,\cite{Martin2020, Malfante2017} 
\begin{align}
    {\rm Kurtosis} \!&=\! 
     \frac{1}{\sigma^2  M}\!\sum_{i=1}^M[y_b[i] \!-\!  \overline{\mathbf{y}_b} ]^4 \! \label{eq:kurtosis}
\end{align}
$\sigma^2$ is the variance and $\overline{\cdots}$ is the arithmetic mean. 

The number of peaks and cross-sensor coherence were extracted from the Hilbert transform $H()$ of the beamformed pressure timeseries (Fig.~\ref{fig:feature_extract})
\begin{align}
   \mathbf{Y}_b = H(\mathbf{y}_b) \nonumber
\end{align}
Here, the number of peaks is the number of local maxima with at least 5 dB prominence relative to the standard deviation, 
\begin{align}\label{eq:npeaks}
    {\rm Npeaks} &= \!\!\sum_{j\in (a,b)}\!\!\mathbb{I}(\max_{j} Y_b[j]>\sigma\cdot 10^{5/10}\!+\!\max\min_j Y_b[j]),
\end{align}
where $(a,b)$ is an interval in ${\mathbf{Y}_b}$, $\max \min_j Y_b[j]$ indicates the highest trough in $(a,b)$, and $\sigma$ is the standard deviation of $Y_b$. 

Last, for the experimental detections, the normalized correlation coefficient of the timeseries envelope across DASARs was computed to measure the spatial coherence of the signal propagation,
\begin{align}
    {\rm Coherence} &= \max_i \frac{1}{C}\sum_{m=i}^{M-1} {Y}_{b}^N[m] {Y}_{b}^S[m\!-\!i]\\
    C & = {\sqrt{\|{\mathbf{Y}}_{b}^N\|_2^2 \!+\!\|{\mathbf{Y}}_{b}^S\|_2^2}},\nonumber
\end{align}
where $\mathbf{Y}_{b}^N$ and $\mathbf{Y}_{b}^S$ are the Hilbert transform of the beamformed timeseries at the North and South DASARs. The measure of coherence assumes that a good quality biological signal will be received as similar timeseries on two spatially separated sensors.

\begin{figure}[tb]
\centering
\includegraphics[width = 0.48\textwidth]{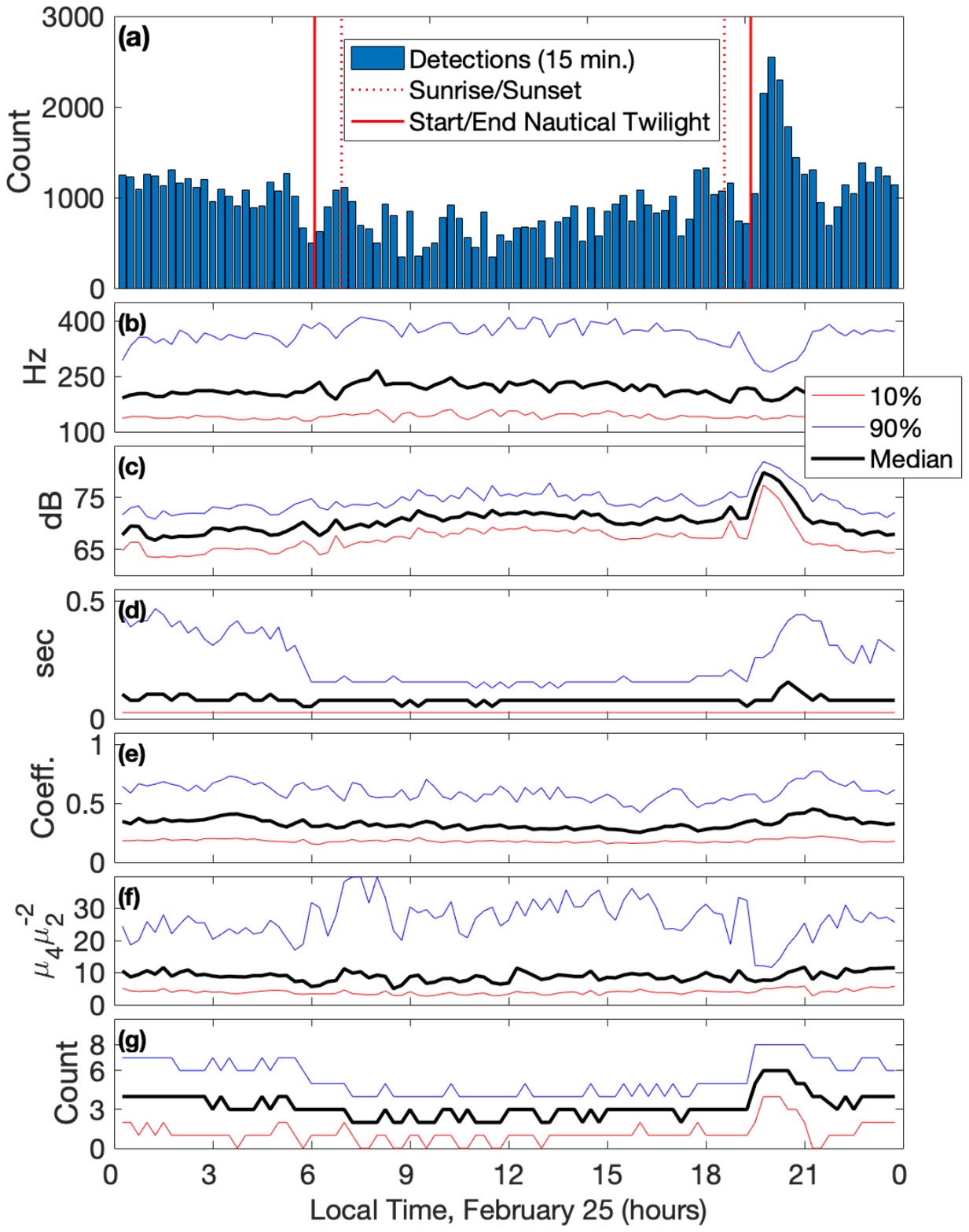}
\caption[{ (a)} Number of detected events per 15 minutes on February 25 and { (b--g)} 10\%, 50\%, and 90\% levels per 15 minutes for each feature: {(b)} peak frequency, { (c)} median time-frequency power, { (d)} event duration, { (e)} normalized time coherence between sensors, { (f)} kurtosis, and { (g)} number of time peaks.]{\label{fig:events_per_hour}{ (a)} Number of detected events per 15 minutes on February 25 and {(b--g)} 10\%, 50\%, and 90\% levels per 15 minutes for each feature: { (b)} peak frequency, {(c)} median time-frequency power, { (d)} event duration, { (e)} normalized time coherence between sensors, {(f)} kurtosis, and {(g)} number of time peaks. All features were measured on the DASAR S. The coherence is the normalized correlation lag coefficient between DASARs N and S.}
\end{figure}

Fig.~\ref{fig:events_per_hour} shows the number of events detected along with extracted feature median, 10\%, and 90\% levels for every 15 minutes. The number of peaks in the timeseries has been shown to be an indicator of fish species and call context for Hawaiian reef fish.\cite{TricasBoyle2014reeffish, Maruksa2007} 
In this study, fish were most common during nighttime (Fig.~\ref{fig:events_per_hour}g), with pulse trains peaking during the evening chorus after nautical twilight (19:15 HST). The evening chorus corresponded to a visible increase in median power, event duration, and number of time peaks and a visible decrease in the $90$th percentiles of peak frequency and kurtosis. 


\section{Results\label{sec:Results}}
\subsection{Simulations\label{sec:SimResults}}
\begin{figure}[tb]
\centering
\includegraphics[width=0.5\textwidth]{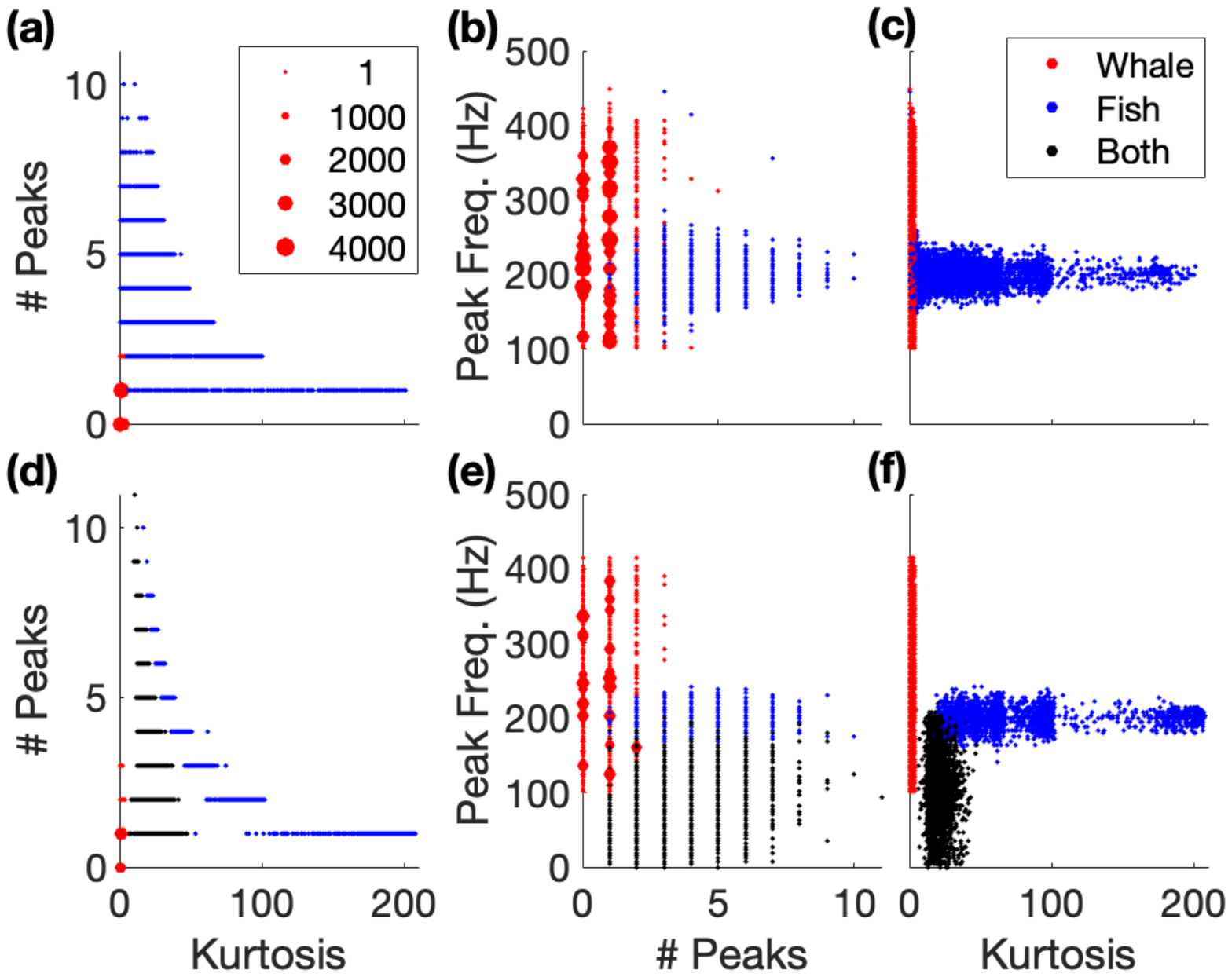}
\caption[Handpicked features demonstrate some separability by kurtosis, number of timeseries peaks, and peak frequency but with overlapping clusters for {(a-c)} two clusters (whale, fish) and { (d-f)} (bottom) three clusters (whale, fish, both). The dots are scaled to indicate density of feature pairs, with each dot increased by 1 pt for every 100 samples.]{\label{fig:pairplot}Handpicked features demonstrate some separability by kurtosis, number of timeseries peaks, and peak frequency but with overlapping clusters for { (a-c)} two clusters (whale, fish) and {(d-f)} (bottom) three clusters (whale, fish, both). The dots are scaled to indicate density of feature pairs, with each dot increased by 1 pt for every 100 samples. 
}
\end{figure}

The handpicked features were examined for separability for $K\!=\!2$ equal-sized clusters (fish, whale) and $K\!=\!3$ clusters (fish, whale, both). Figure~\ref{fig:pairplot} shows the feature values colored by true signal type for one simulation trial. The whale and fish signals overlapped in their peak frequency and in 
the number of automatically extracted temporal peaks. The signals were most strongly separated in kurtosis, with whale signals having very low kurtosis. 

For 2 clusters, all clustering methods assumed $K\!=\!2$. When the cluster sizes were equal, all clustering methods achieved high accuracy (Table~\ref{tab:results1}). 
For the unequal cluster scenario 
with 25\%  whale and 
75\% fish (2500/7500), 
GMM achieved much higher accuracy than either K-means or hierarchical agglomerative clustering on the handpicked features (Table~\ref{tab:results1}). These simulation results are supported by a supplementary example of the clustering algorithm assumptions (App.~\ref{appendix}), which demonstrates that GMM has fewer prior assumptions and is thus more flexible. 
The classification accuracy of all methods for 3 equal-sized clusters (Table~\ref{tab:results1}), with whale, fish, and a cluster containing a temporally overlapping whale and fish, was lower than for 2 equal-sized clusters due to the signal overlap. GMM is capable of estimating overlapping clusters and achieved the best accuracy and recall on the handpicked features of the overlapping signal scenario. 


\begin{table*}[tb]
\footnotesize
\caption{\label{tab:results1} Median of classification accuracy, recall, and precision across 100 simulations for clustering with hand-crafted features (K-means, GMM, hierarchical) and deep learning (DEC). 
 Whale was defined as the positive category.}
\begin{center}
\begin{tabular}{l c @{\hspace{\tabcolsep}} c @{\hspace{\tabcolsep}}c  p{0.1mm}   c@{\hspace{\tabcolsep}} c@{\hspace{\tabcolsep}} c  p{0.1mm}    c@{\hspace{\tabcolsep}} c@{\hspace{\tabcolsep}} c@{\hspace{\tabcolsep}}}
\hline\hline
Method & Accuracy  & Recall & Precision &  & Accuracy  & Recall & Precision &  &  Accuracy  & Recall & Precision  \\
\hline
&
\multicolumn{3}{c}{{$K\!=\!2$ simulated clusters}} 
&

&
\multicolumn{3}{c}{{$K\!=\!3$ simulated clusters}}
&

&
\multicolumn{3}{c}{{$K\!=\!2$ (unequal)}}
\\
 & \multicolumn{3}{c}{{(DEC $P\!=\!10$)}} & & \multicolumn{3}{c}{{(DEC $P\!=\!15$)}} & & \multicolumn{3}{c}{{simulated clusters}}\\
 & &  & & & & & & & \multicolumn{3}{c}{{(DEC $P\!=\!10$)}} 
 \\

K-means & {\bf 0.96}
& \hspace{3pt} 1.0 \hspace{3pt} 
& 0.93 
 
&
 &  {\bf 0.56}
&  0.65 
& 0.99 

&
&  {\bf 0.68}
&  \hspace{3pt} 1.0 \hspace{3pt}
&  0.44
\\

GMM & {\bf 0.98}
& \hspace{3pt} 1.0 \hspace{3pt}
&  0.96
 
&
& {\bf 0.86}
& \hspace{3pt} 1.0 \hspace{3pt} 
& 0.94 

 &
  & {\bf 0.97}
& \hspace{3pt} 1.0 \hspace{3pt} 
&  0.88 
\\
Hierarchical  & {\bf 0.98}
& \hspace{3pt}0.99 
&  0.98 

&
&  {\bf 0.60}
& 0.58
& 0.99

&
& {\bf 0.66}
 & \hspace{3pt}1.0
 &  0.42 
 \\
 GMM, {\scriptsize latent features} & {\bf 0.95}
& 0.96 
& 0.99 

&
&{\bf 0.69}
& 0.90
& \hspace{3pt} 1.0 \hspace{3pt} 

&
& {\bf 0.96}
& 0.89 
& \hspace{3pt} 1.0 \hspace{3pt} 
 \\ 
DEC & {\bf 0.98}
& 0.99 
& 0.99 

&
& {\bf 0.75} 
& 0.92
& 0.99

&
& {\bf 0.77}
 & \hspace{3pt} 1.0 \hspace{3pt} 
 & 0.52 
 \\
\hline\hline
\end{tabular}
\end{center}
\end{table*}

\begin{figure}[tb]
\centering
\includegraphics[width=0.5\textwidth]{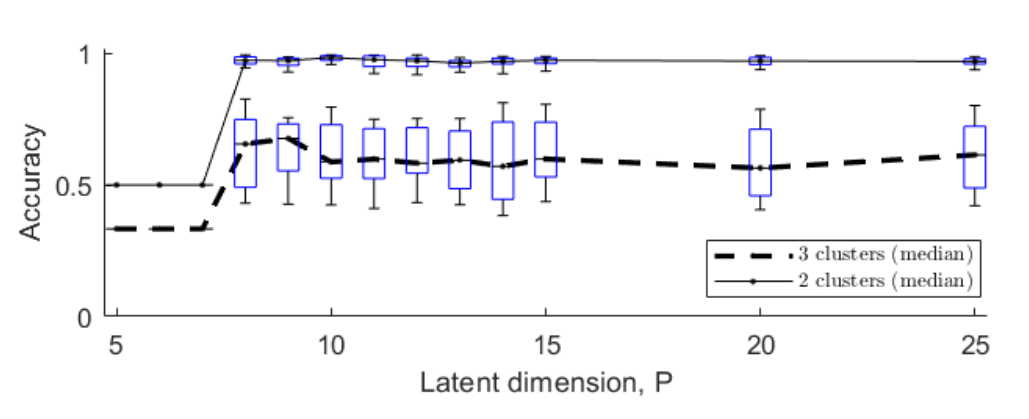}
\caption[Accuracy of DEC over 100 trials for each latent dimension ($P$). The whiskers show the $10$th (lower) and $90$th (upper) percentile. $K\!=\!3$ clusters contained fish, whale, or fish and whale signals, and $K\!=\!2$ clusters had fish or whale.]{\label{fig:hyperparameter} Accuracy of DEC over 100 trials for each latent dimension ($P$). The whiskers extend to the $10$th (lower) and $90$th (upper) percentiles. The box marks the $25$th to $75$th percentile. 
The method fails at low $P$ but is otherwise stable.}
\end{figure}

The effect of the DEC latent feature dimension was examined across 100 random simulation trials (Fig.~\ref{fig:hyperparameter}). The DEC model, motivated by a successful architecture for classifying seismic spectrograms, 
\cite{Snover} 
was retrained and tested on all trials at varying latent feature vector dimensionality ($P$) for the 
$K\!=\!2$ equal-sized clusters and $K\!=\!3$ cluster scenarios 
 (Fig.~\ref{fig:hyperparameter}). 
When the latent dimension was low, {\it e.g.} $P\!<\!8$, all features were likely to be clustered in a single cluster. 
Above $P\!>\!8$, DEC accuracy was consistent. Values of $P\!=\!10$ and $P\!=\!15$ were chosen for the $K\!=\!2$ and $K\!=\!3$ cluster scenarios. 

\begin{figure}[tb]
\centering
\includegraphics[width=0.5\textwidth]{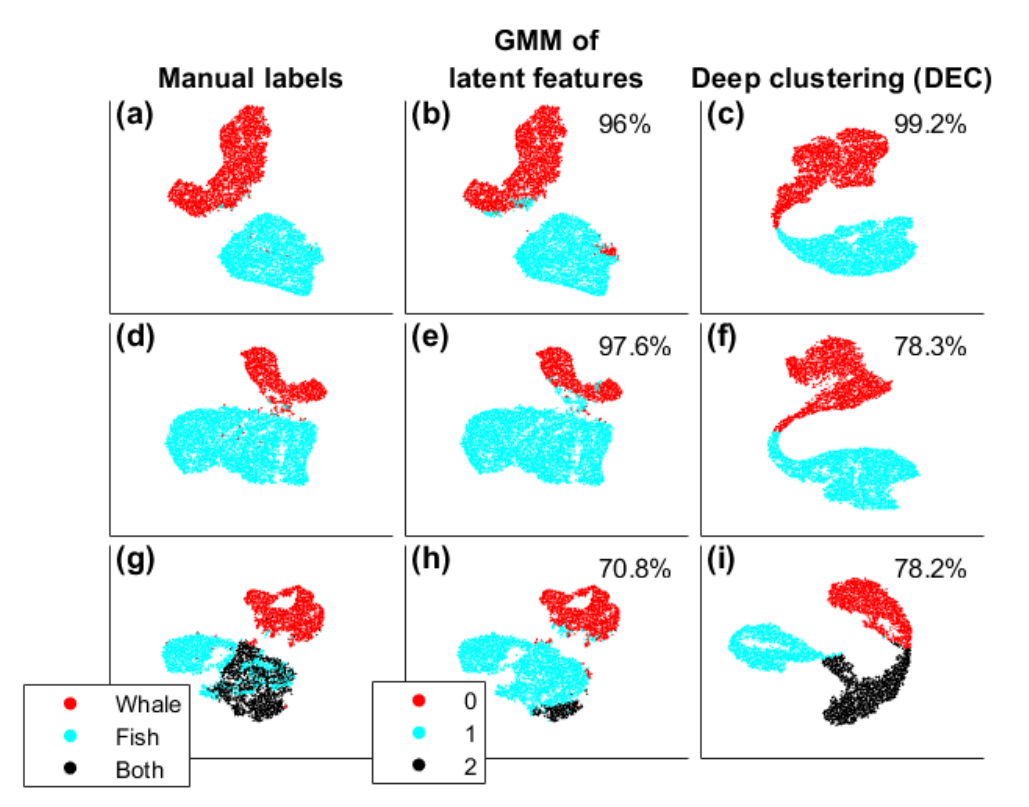}
\caption{\label{fig:DEC_sim} t-SNE representation of 10000 deep embedded feature vectors from a sample trial {(a,b,d,e,g,h)} before, and (c,f,i) after deep clustering, with 
(a-c) equal-sized clusters, {(d-f)} unequal-sized clusters of fish and whale, and {(g-i)} fish, whale, and fish/whale clusters. 
The perplexity for visualization was 200.}
\end{figure}

Figure~\ref{fig:DEC_sim} shows t-SNE visualizations from a representative trial of the latent features  clustered with GMM (Fig.~\ref{fig:DEC_sim}(b,e,h)), and the DEC latent features and cluster predictions after 20 additional training epochs using \eqref{eq:KL} (Fig.~\ref{fig:DEC_sim}(c,f,i)). 
Additional training with \eqref{eq:KL} increased the accuracy of DEC clustering prediction from 96\% to 99.2\% in the case of 2 equal sized clusters and from 70.8\% to 78.2\% in the case of 3 clusters. In the scenario containing 2 unequal-sized clusters, the additional DEC training reduced the accuracy from 97.6\% to 78.3\%. In that case, high recall and low precision (Table~\ref{tab:results1}) indicate that DEC overpredicted the smaller cluster. GMM clustering of the latent features learned from spectrograms achieved comparable accuracy to GMM clustering of handpicked features in the non-overlapping scenarios with $K\!=\!2$ (Table~\ref{tab:results1}).
%
\begin{figure}[tb]
\centering
\includegraphics[width=0.5\textwidth]{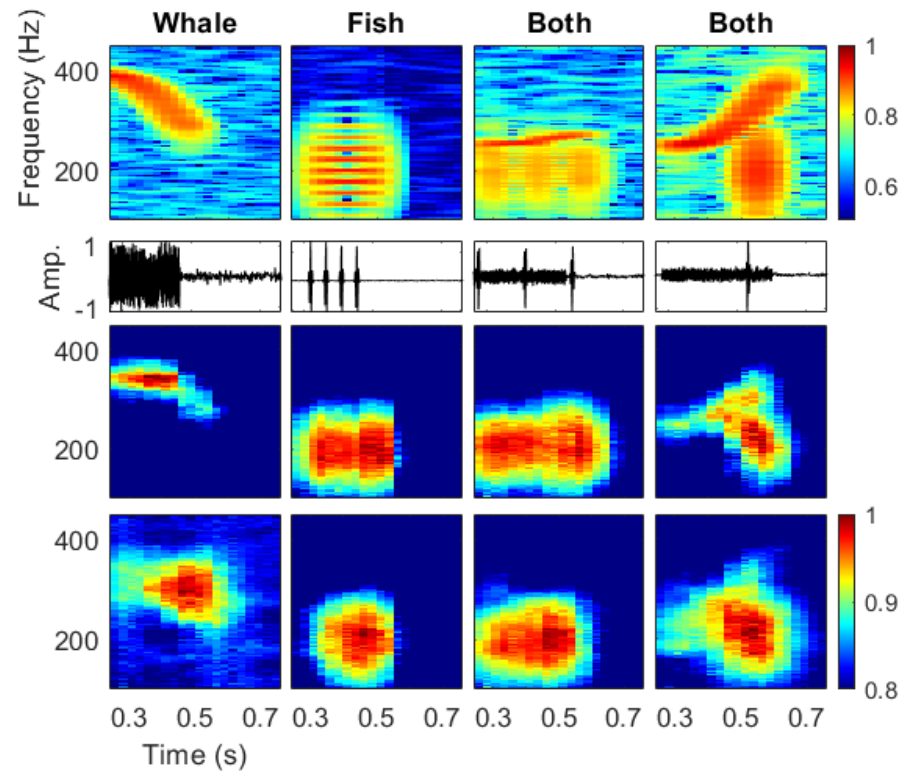}
\caption[Simulated coral reef bioacoustic signals successfully classified using DEC. From top to bottom rows: input spectrograms, simulated timeseries, initial DEC reconstruction, and DEC reconstruction after adding clustering loss.]{\label{fig:sim_examples} Simulated coral reef bioacoustic signals successfully classified using DEC. From top to bottom rows: input spectrograms, simulated timeseries, initial DEC reconstruction, and DEC reconstruction after adding clustering loss.}
\end{figure}
For the $K\!=\!3$ scenario, the highest classification confusion for the spectrogram-based deep learning was between fish and the overlapping signal cluster. Figure~\ref{fig:sim_examples} shows that fish signals with large bandwidth and duration may dominate the spectral signature, which may lead to increased misclassification.

\subsection{Experiment}\label{sec:RealResults}
The clustering methods were applied to a subset of 4,000 unlabeled detections randomly selected from the Hawaii 2020 experiment. Each detection contained one or more directional signals. All clustering methods assumed $K\!=\!2$.  
Then, for post-clustering analysis, labels of whale/no whale (only fish) were manually assigned to all 4000 samples, based on the signal within the detection window. About two-thirds were labeled as no whale and contained only fish. 

\begin{figure}[tb]
\centering
\includegraphics[width=0.35\textwidth]{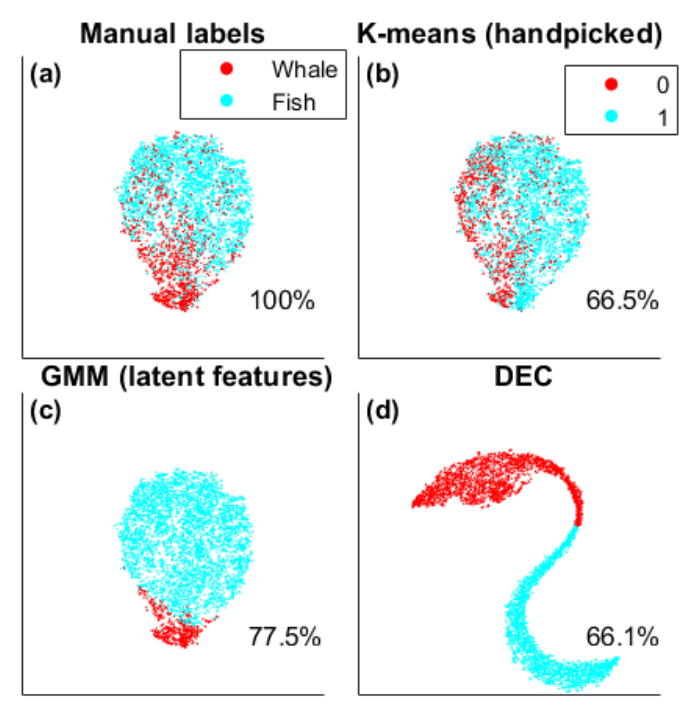}
\caption{\label{fig:tsne_exp} Features of experimentally detected signals from a Hawaiian coral reef shown as a t-SNE representation of {(a--c)} learned latent features and 
{(d)} modified DEC features of 4,000 manually labeled detections. All clustering methods assumed $K\!=\!2$.}
\end{figure}

Figure~\ref{fig:tsne_exp} shows the clustering results using t-SNE representations of the learned latent feature vectors (Fig.~\ref{fig:tsne_exp}(a-c)) and the DEC latent features after additional training (Fig.~\ref{fig:tsne_exp}(d)). Clustering of the $P\!=\!6$--dimensional handpicked feature vectors did not correspond well to the manual labels, with K-means achieving the highest accuracy of 67\%. GMM clustering of the latent features learned from the signal spectrograms achieved an accuracy of 78\% compared to the manual labels, which was the highest accuracy of all methods examined. GMM of latent features had 
high precision and low recall (Table~\ref{tab:results2}).

In the experimental signals, additional training of the DEC latent features reduced the classification accuracy (Fig.~\ref{fig:tsne_exp}(d)). Similar to the simulated unequal-sized cluster scenario, DEC obtained higher recall with lower precision (Table~\ref{tab:results2}). 

\begin{table}[tb]
\caption{\label{tab:results2} Classification accuracy determined from manually labeled experimental detections for clustering with handpicked features (K-means, GMM, hierarchical) and deep learning (DEC). }
\begin{center}
\begin{tabular}{c  c c c}
\hline\hline
Method & Accuracy & Recall & Precision\\
\hline
K-means ($K\!=\!2$) & {\bf 0.66} & 0.42  &  0.47 \\
GMM ($K\!=\!2$) & {\bf 0.55} & 0.84 & 0.40 \\
Hierarchical ($K\!=\!2$) & {\bf 0.63}  & 0.23 & 0.38  \\
GMM ($K\!=\!2$), latent features & {\bf 0.78} & 0.33 & 0.90\\
DEC ($P\!=\!15$, $K\!=\!2$) & {\bf 0.66} &  0.75 & 0.48\\
\hline\hline
\end{tabular}
\end{center}
\end{table}


\begin{figure}[tb]
\centering
\includegraphics[width=0.5\textwidth]{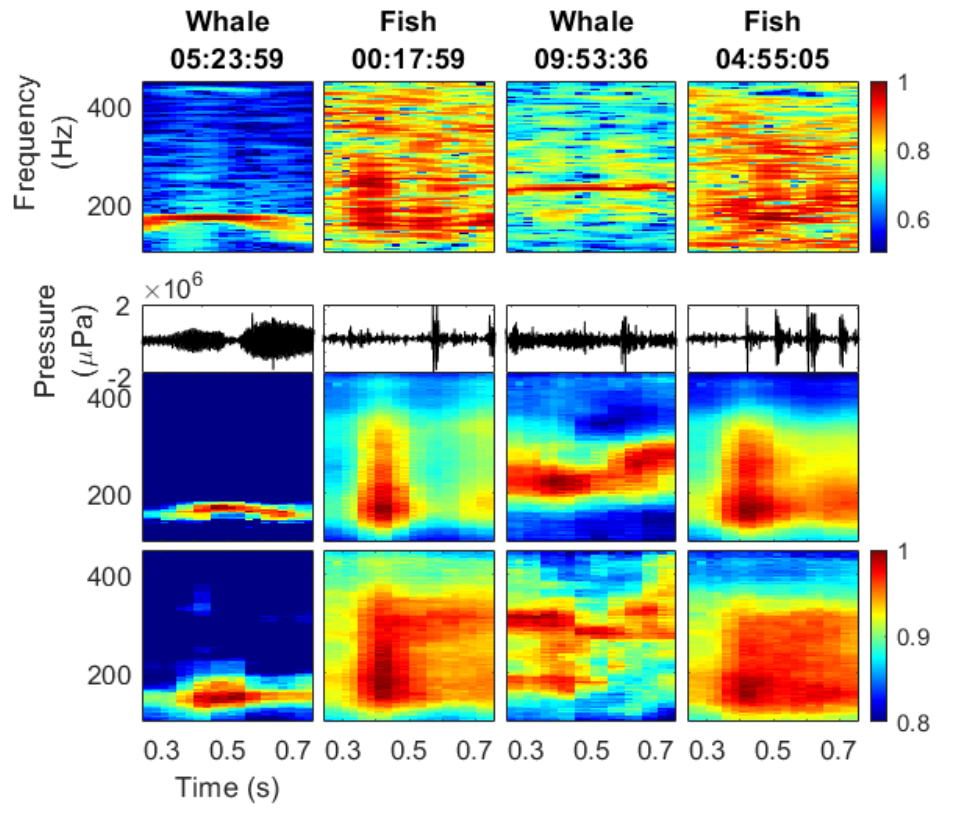}
\caption[Experimentally detected coral reef bioacoustic signals successfully classified using DEC.]{\label{fig:correct_eg} Experimentally detected coral reef bioacoustic signals successfully classified using DEC. From top to bottom rows: input spectrograms, recorded timeseries, initial DEC reconstruction, and DEC reconstruction after adding clustering loss. 
}
\end{figure}

Spectrograms correctly classified by DEC are shown in Fig.~\ref{fig:correct_eg}, with the reconstructed spectrograms shown before and after additional DEC training using \eqref{eq:KL}. These demonstrate that whale signals were primarily identified by their narrow bandwidth and temporal extent, whereas fish signals were identified as broadband. The timeseries in Fig.~\ref{fig:correct_eg} demonstrate the magnitude variation between events that was not attributed to signal type, motivating the normalization of the spectrograms.

\section{Conclusion\label{sec:discussion}}

A deep clustering approach was presented for interpreting unlabeled coral reef bioacoustic detections. This approach leverages deep learning and clustering, motivated by recent improvements in classification accuracy of fish calling using unsupervised detection and deep neural network classifiers.\cite{Malfante2017, Lin2018, Ibrahim2018, Ibrahim2019_SAE}


Clustering of simulated fish (Gaussian pulses) and whale (FM sweeps) demonstrated that unique signal categories could be clustered with GMM of deep latent features learned from spectrograms, DEC clustering,\cite{Snover, GuoDEC} or with clustering of handpicked features motivated by studies of coral reef fish.\cite{MannLobel1997, Maruksa2007, TricasBoyle2014reeffish} GMM applied to either deep latent features or handpicked features achieved the highest accuracy when there were two clusters of unequal size, and GMM of handpicked features had the highest accuracy when the clusters contained overlapping signals. 
Deep clustering learned features directly from spectrograms and was successful in separating fish and whale clusters. DEC trained with a joint clustering loss function increased the classification accuracy in simulations compared to clustering the deep latent features with GMM, except when the cluster sizes were significantly different. For unequal-sized simulated clusters, applying GMM to the learned latent features demonstrated higher classification accuracy. 

Broadband bioacoustic events detected on a Hawaiian coral reef in February--March 2020 using a directional detector 
were analyzed with deep clustering and clustering of handpicked features. A labeled subset of these detections with whale/no whale indicated that about two-thirds of the detections contained primarily fish and one-third contained a whale signal. 
GMM with $K\!=\!2$ applied to latent features learned from the signal spectrograms 
achieved the highest accuracy and tended to overpredict the larger cluster containing fish. After further training DEC with a joint clustering loss, the DEC recall increased but DEC classification accuracy was reduced, indicating that  
whale was overpredicted. 
DEC reconstructions of the input spectrograms demonstrate that the learned features are representative of spectral features identified by manual labelers. Clustering 
of handpicked features with $K\!=\!2$ on the experimental, manually labeled data achieved lower accuracy, precision, and recall. 

These results demonstrate that deep clustering is a promising method for classifying unlabeled bioacoustic signals with distinct spectral signatures. In simulated and experimental studies, both GMM applied to deep latent features and DEC trained with a joint clustering loss achieved competitive classification accuracy for a range of latent feature dimensions, without requiring the selection of handpicked features. Our results indicate that the accuracy of handpicked feature clustering depends strongly on the 
feature properties and the choice of clustering method. 
Finally, the choice of clustering algorithm is an important consideration for applications that are subject to signal clusters of unequal size.

\section{Acknowledgements} 
Thank you to Greeneridge Sciences for providing the DASAR sensors and to Alex Conrad for assisting with DASAR post-processing. Thank you to Richard Walsh for assistance with hardware deployment and recovery. This work was supported by the Office of Naval research under award N00014-18-1-2065. 

\appendix

\section{Clustering of Gaussians}\label{appendix}
For a continuous data variable,
$\mathbf{x}$, that is
Gaussian with 
mean $\bm{\mu}_k$ 
and covariance $\bm{\Sigma}_k$, 
the posterior probability of the cluster is given by,\cite{Bishop}
\begin{align}
     p(C_k | \mathbf{x}) &= \frac{p(\mathbf{x}|C_k)p(C_k)}{\sum_j p(\mathbf{x}|C_j)p(C_j)}\\
    p(\mathbf{x} | C_k) \!&=\! \frac{1}{(2\pi)^{\frac{P}{2}}|\bm{\Sigma}_k|^{\frac{1}{2}}}  e^{ -\frac{1}{2}(\mathbf{x} \!-\! \bm{\mu}_k)^T\bm{\Sigma}_k^{-1} (\mathbf{x} \!-\! \bm{\mu}_k)},\label{eq:Gaussian_likelihood}
\end{align}
where $C_k$ and $C_j$ represent the $k$th and $j$th classes and $|\bm{\Sigma}_k|$ is the determinant of $\bm{\Sigma}_k$. 
The boundary between the two classes occurs when there is an equal probability of $\mathbf{x}$ belonging to either class, 
\begin{align}
   &\quad\quad\log p(C_k|\mathbf{x}) = \log p(C_j | \mathbf{x})\label{eq:class_equality} \\
     0=& \hspace{-0.7mm}-\mathbf{x}^T (\bm{\Sigma}_k^{-1}\hspace{-1mm}-\!\!\bm{\Sigma}_j^{-1})^{ } \mathbf{x} \hspace{-1mm}+\!\! 2(\bm{\Sigma}_k^{-1}\!\bm{\mu}_k\!\!\!-\!\!\!\! \bm{\Sigma}_j^{-1}\!\bm{\mu}_j)^T\mathbf{x} +  C\!\!\nonumber\\
    C = &\!-  \bm{\mu}_k^T \bm{\Sigma}_k^{-1} \bm{\mu}_k \!\!+\!\! \bm{\mu}_j^T \bm{\Sigma}_j^{-1}\bm{\mu}_j \!\!-\!\!\log\frac{|\bm{\Sigma_{\normalfont k}}|}{|\bm{\Sigma}_j|}\! \!+\!\! 2\log \frac{p(C_j)}{p(C_k)}\!\!\nonumber
\end{align}
The general solution for $\mathbf{x}$ in \eqref{eq:class_equality} is a multi-dimensional parabola, which simplifies to a linear boundary if the distributions have a shared covariance such that $\bm{\Sigma}_k = \bm{\Sigma}_j$ $\forall j,k$. The solution in \eqref{eq:class_equality} is extensible to $K>2$ classes by considering the joint distributions of all classes (see Bishop Ch.~4.2.1\cite{Bishop} for details).

\subsection{Clustering simulations}
\begin{figure}[tb]
\centering
\includegraphics[width = 0.45\textwidth]{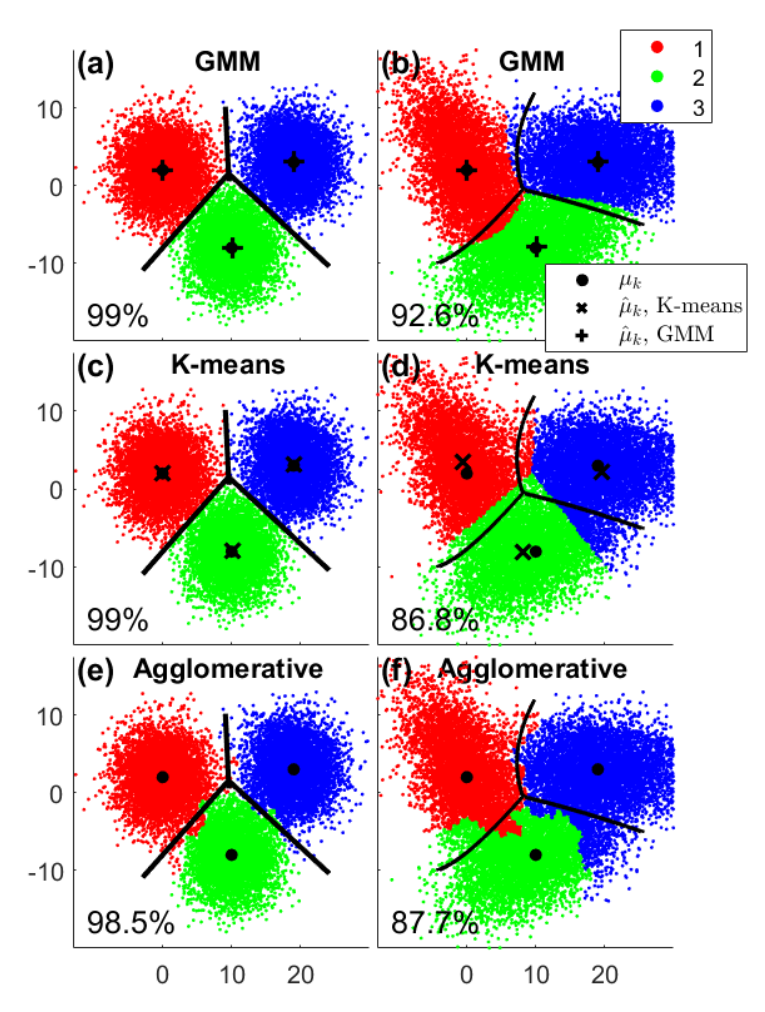}
\caption[Clustering on two data distributions using {(a,b)} GMM, {(c,d)} K-means, and { (e,f)} agglomerative clustering with Ward's method, with maximum likelihood boundaries shown by black lines.]{\label{fig:distances} Clustering on two data distributions using {(a,b)} GMM, { (c,d)} K-means, and { (e,f)} agglomerative clustering with Ward's method, with maximum likelihood boundaries shown by black lines. }
\end{figure}
Three 2D Gaussian distributions, each with $N=10,000$ points, were used to simulate overlapping clusters (Fig.~\ref{fig:distances}). The true cluster means were $\bm{\mu}_1 = (0,2)$, $\bm{\mu}_2=(10, $-$8)$, and $\bm{\mu}_3 = (21,3)$. The covariance of the first dataset was
\begin{align}
    \Sigma_{\mathcal{C}_1} = \begin{bmatrix}
   \sigma_x^2 & 0 \\ 
   0 & \sigma_y^2 
   \end{bmatrix}
\end{align}
with $\sigma_x^2 =\sigma_y^2= 3$. For the first dataset (Fig.~\ref{fig:distances}a), there were no off-diagonal covariance terms.

The second dataset (Fig.~\ref{fig:distances}b) was generated by rotating the data counterclockwise at $\theta$, with
\begin{align}
    \Sigma_{\mathcal{C}_2} \!=\!  \begin{bmatrix}
   \cos(\theta) &\!\! -\!\sin(\theta) \\
   \sin(\theta) &\!\! \cos(\theta)
   \end{bmatrix}\!\!
   \begin{bmatrix}
   \sigma_x^2 & 0 \\ 
   0 & \sigma_y^2 
   \end{bmatrix} \!\!
   \begin{bmatrix}
   \cos(\theta) & \!\!-\!\!\sin(\theta) \\
   \sin(\theta) &\!\! \cos(\theta)
   \end{bmatrix}^T
\end{align}
with $\sigma_x^2 = 6$, $\sigma_y^2$ = 3. The three clusters were rotated by $\theta=120^\circ, 25^\circ,$ and $0^\circ$. 
The non-isotropic covariance is a result of correlation between variables and indicates feature correlations. 
A non-isotropic covariance matrix of a feature vector indicates correlation between features.

GMM assumes that the clusters are Gaussian distributed around a cluster mean. 
The K-means algorithm further assumes the Gaussians have the same diagonal covariance and uniform prior probability. Thus, GMM and K-means perform best for the first dataset (Fig.~\ref{fig:distances}(a,c,e)) with identical, diagonal covariance for all clusters. 
By incorporating intercluster distance through the Ward metric, hierarchical agglomerative identifies 3 similar clusters.

When the cluster covariances are not of the form $\sigma^2\mathbb{I}$, K-means and agglomerative clustering provide solutions that are suboptimal to the true class boundaries. GMM achieves better accuracy (Fig.~\ref{fig:distances}(b,d,f)) but does not converge to the maximum likelihood solution due to estimation errors in the covariances. All three methods provide insight into the cluster memberships despite invalid class assumptions, but GMM is the most versatile clustering method. 

\subsection{K-means++ algorithm}\label{app:k++}
K-means++ is an algorithm proposed to efficiently initialize parameters for Gaussian clustering methods, including GMM and K-means. K-means++ improves the runtime of the clustering algorithms and the quality of the final solution.\cite{k++}
For a set of data samples $\cal{X}=\{\mathbf{x}_{\rm 1},\dots,\mathbf{x}_{\it N}\}$, the K-means++ algorithm follows:\cite{k++}
\begin{enumerate}
    \item Initialize the priors $\pi_k=\frac{1}{K}$, $k=1,\dots,K$.
    \item Initialize the covariance matrices as $\bm{\Sigma}=\bm{\sigma}^2\mathbb{I}$, with $\bm{\sigma}^2=$var$[\cal{X}]$.
    \item Select the first cluster mean as a random data sample, $\bm{\mu}_{\it \rm 1}=\mathbf{x}_l$, $l\in \cal{U}({\rm 1},{\rm N})$.
    \item Assign all data samples to the cluster with the nearest mean. We denote that the $m$th sample belongs to cluster $C_i$ as $\mathbf{x}_m\!\in\! C_i$, satisfying $d(\mathbf{x}_m, \bm{\mu}_i)\leq  d(\mathbf{x}_m, \bm{\mu}_p)$, $i,p\!=\!1,\dots,j$, $j\leq K$. Note that when $j=1$ (first step), $\mathbf{x}_m \in C_1$ $\forall m$, {\it i.e.} all data samples belong to one cluster.
    \item Select the $j\!\!+\!\!1$ th cluster mean at random from the remaining data samples with probability \begin{align}\label{eq:pk++}
        p(\mathbf{x}_n\!=\!\bm{\mu}_{j+1}) \!&=\! \frac{d^2(\mathbf{x}_n, \bm{\mu}_{j+1})}{\sum_{\mathbf{x}_m\in C_i} d^2(\mathbf{x}_m,\bm{\mu}_i)}, \quad \!\!\! n\!=\!1,\dots,N\!-\!j, n\neq l
    \end{align}
$d$ is the Mahalanobis distance,
    \begin{align}
        d(\mathbf{x}_i,\mathbf{x}_j) &= \sqrt{(\mathbf{x}_i\!-\!\mathbf{x}_j)^T\bm{\Sigma^{-1}}(\mathbf{x}_i\!-\!\mathbf{x}_j)}.
    \end{align}
    \eqref{eq:pk++} ensures that each mean is selected with a probability proportional to its distance from all existing means.
    \item Repeat Steps 4 and 5 until $K$ cluster means are chosen. 
\end{enumerate}
The initialized parameters define $\bm{\theta}^0=(\pi_k, \bm{\mu}_k, \bm{\Sigma}_k)$, $k=1,\dots,K$ for the first E-step of the GMM algorithm in \eqref{eq:Estep}. For K-means, the prior and covariance are assumed fixed, and the means are updated according to \eqref{eq:kmeans_update}.


\subsection{Visualization of high-dimensional data}\label{sec:tsne}
 For data with more than two dimensions, $\mathbf{x}_n\in\mathbb{R}^P$ for $P\!>\!2$, clusters may be visualized by applying dimensionality reduction. In this study, 2D t-Stochastic Neighbor Embedding (t-SNE)\cite{tsne} was used to visualize $P$--dimensional features.

 The similarity of one point, $\mathbf{x}_i\in \mathbb{R}^P$, to another point, $\mathbf{x}_j\in\mathbb{R}^{P}$, is found from the conditional probability that the points are neighbors within a Gaussian density with mean $\mathbf{x}_i$,\cite{tsne}
\begin{align}\label{eq:symmetric_probs}
    p_{j|i} &= \frac{e^{-\frac{1}{2\sigma_i^2}\|\mathbf{x}_i - \mathbf{x}_j\|^2} }{\sum_{k\neq l} e^{-\frac{1}{2\sigma_i^2}\|\mathbf{x}_k - \mathbf{x}_l\|^2}},\quad
    p_{i|i} = 0
    \end{align}
for $N$ points $i,j$=$1,\dots,N$. 
\begin{figure}[tb]
\centering
\includegraphics[width=0.5\textwidth]{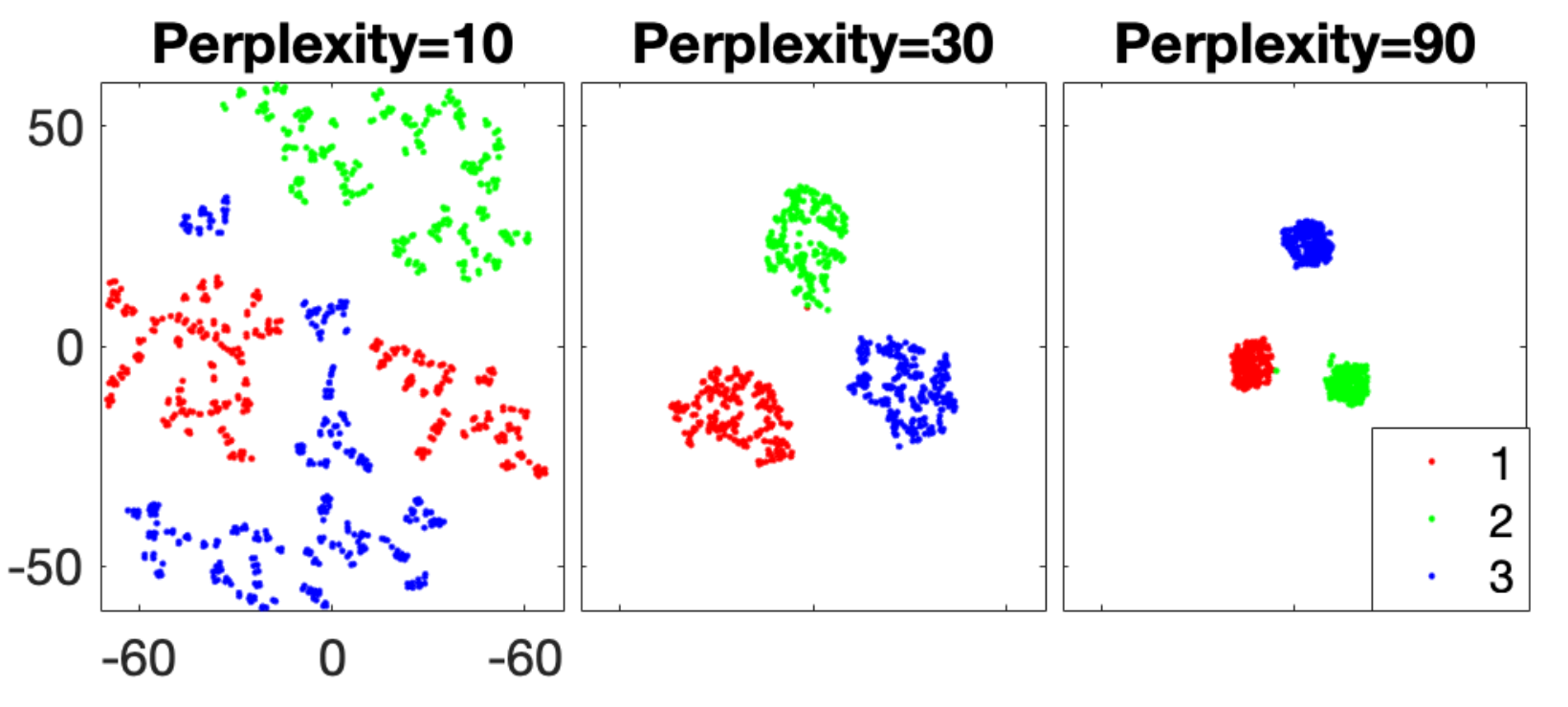}
\caption[Varying values of perplexity for t-SNE on $N\!=\!1000$ randomly drawn points $\mathbf{x}_n\in\mathbb{R}^3$, $n=1,\dots,N$.]{\label{fig:perplexity} Varying values of perplexity for t-SNE on $N\!=\!1000$ randomly drawn points $\mathbf{x}_n\in\mathbb{R}^3$, $n=1,\dots,N$. }
\end{figure}
The neighborhood of $\mathbf{x}_i$, as determined by $\sigma_i$, 
is defined implicitly in terms of the perplexity (Fig.~\ref{fig:perplexity}),\cite{tsne, BarnesHutSNE}
\begin{align}\label{eq:perplexity}
    \text{perplexity}(P_i) &= 2^{H(P_i)}\\
    H(P_i) &= -\sum_{j=1}^N p_{j|i} \log_2 p_{j|i}
\end{align}
where $P_i =\sum_j p_{j|i}$, and $H$ is the Shannon entropy. The optimal value of $\sigma_i$ in \eqref{eq:symmetric_probs} for each point is solved with a binary search for a given value of perplexity.\cite{BarnesHutSNE} 

Then, a set of projected data is randomly initialized with zero-mean Gaussians of low variance,\cite{tsne} $\mathbf{y}_i\in\mathcal{N}(\mathbf{0}, 10^{-4} \bm{\mathbb{I}})$, $\mathbf{y}_i\in\mathbb{R}^2$. The $\mathbf{y}_i$ are iteratively updated until the Kullback-Leibler (KL) divergence between the conditional probability in \eqref{eq:symmetric_probs} and the Student's t-distribution of the projected data is minimized,
\begin{align}
    q_{j|i} \!=\! \frac{(1+\|\mathbf{y}_i - \mathbf{y}_j\|^2)^{-1} }{\sum_{k\neq l} (1+\|\mathbf{y}_k - \mathbf{y}_l\|^2)^{-1}},\quad q_{i|i}=0
\\
    KL(P\|Q)\!=\! -\sum_{i \neq j}\sum_{j=1}^N p_{j|i} \log \frac{p_{j|i}}{q_{j|i}}.\label{eq:KLdivergence}
\end{align}
In contrast to classic SNE which uses only Gaussians, t-SNE's use of the Student t-distribution further penalizes outliers in the projected data,\cite{tsne} resulting in a more compact representation.
As shown in Fig.~\ref{fig:perplexity}, the value of perplexity should be varied according to user preference to obtain the desired visualization. 

\section{Directional detector}\label{app:detector}
A directional event detector\cite{Thodeinreview} was developed to utilize the DASARs' directional capability by combining two DASARs. The complex spectrograms of the $x$-- and $y$-- particle velocity, matrices $\mathbf{V}_x\in\mathbb{C}^{N_f\times N_t}$ and $\mathbf{V}_y\in\mathbb{C}^{N_f\times N_t}$, were generated identically to the spectrogram $\mathbf{X}$ with units m$\cdot$ s$^{-1}$. 

The active intensity, a measure of in-plane energy, was used to determine the noise directionality
\begin{align}\label{eq:az}
       \mathbf{A} &=  \operatorname{atan2}\left(\Re\{ \mathbf{X}\odot \mathbf{V}_y^*\},\Re \{ \mathbf{X}\odot \mathbf{V}_x^*\}\right), 
\end{align}
$^*$ is the complex conjugate and $\Re$ the real component. Atan2 is a piecewise function that computes the elementwise angle between the elements of two matrices, with domain $(0^\circ, 359^\circ)$ defined counterclockwise from the $y$-axis ($0^\circ$ = North). 
%
$\mathbf{A}$ is therefore the time-frequency representation of compass directionality. In the following, matrix $\mathbf{A}_{\rm N}$ is called the {\it azigram}\cite{Thode_azigram} for DASAR N, likewise for $\mathbf{A}_{\rm W}$ and $\mathbf{A}_{\rm S}$.
 
\begin{figure}[tb]
\centering
\includegraphics[width = 0.48\textwidth]{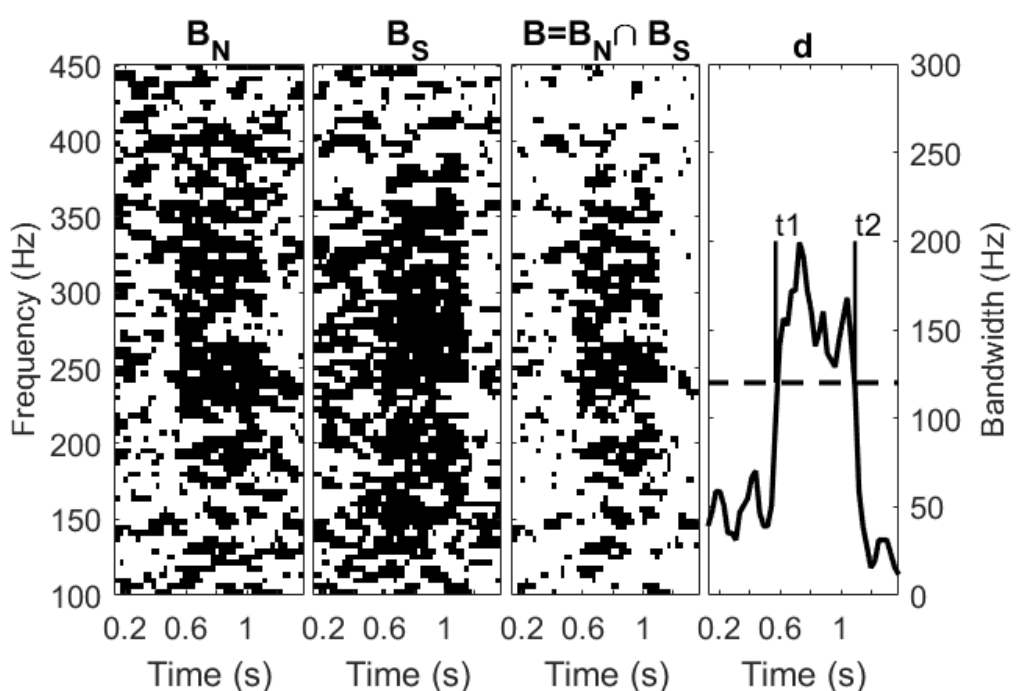}
\caption{\label{fig:azi_detector} Directional detector for a fish call event on February 25, 07:12 HST, with DASAR N looking between 135$^\circ$--225$^\circ$ and DASAR S looking between 45$^\circ$--135$^\circ$ (clockwise from north). The overlap of the binary masks, summed across frequency, defines the detection timeseries.} 
\end{figure}
The event detector makes the following signal assumptions:
 \begin{enumerate}
     \item An event arrives from a constant azimuthal sector for each DASAR. 
     \item Target events are broadband below 500 Hz. The minimum required bandwidth was set with an empirical threshold (see Appendix in Thode {\it et al.} 2021).\cite{Thodeinreview}
 \end{enumerate}
The detection algorithm is demonstrated in Fig.~\ref{fig:azi_detector}. First, the azigrams for the north- and southmost DASARs, $\mathbf{A}_N$ and $\mathbf{A}_S$, were used to create binary maps $\mathbf{B}_N$ and $\mathbf{B}_S$ of time-frequency points within a fixed azimuthal sector (Fig.~\ref{fig:azi_detector}a),

\begin{align}\label{eq:BZ}
    \mathbf{B}_{\rm N}=\mathbb{I}(\pmb{\theta}_{\rm N}^{(1)} < \mathbf{A}_{\rm N} \leq \pmb{\theta}_{\rm N}^{(2)}),
\end{align}
and likewise for $\mathbf{B}_S$ (Fig.~\ref{fig:azi_detector}b). $\mathbb{I}$ is the elementwise indicator function, with $\mathbb{I}(\mathbf{true}) = \mathbf{1}$. Binary maps were generated for all combinations of azimuthal sectors $\bm{\theta}_{\rm N},$ $\bm{\theta}_{\rm S} \in ([0, \Delta\theta]^T, [\frac{\Delta\theta}{2}, \frac{3\Delta\theta}{2}]^T, \dots,[(360^\circ\!-\!\Delta\theta), 360^\circ])$. Here, $\Delta\theta = 90^\circ$.
Next, overlapping signals on both DASARs were discovered by creating a combined map (Fig.~\ref{fig:azi_detector}c), which was summed across frequency to determine the detection timeseries,
\begin{align}\label{eq:B}
    \mathbf{B} = \mathbf{B}_{\rm N} \cap \mathbf{B}_{\rm S},\\
    \mathbf{d} = df*\sum_i \mathbf{B}(i,:)\label{eq:d}
\end{align}
$\mathbf{d}$ measures the bandwidth of an event. Event start and end times were determined for $d_j>T$, $j=1,\dots,N_t$ for threshold $T$.  
Events separated by less than $M_{\rm sep}\cdot dt$ were merged and events longer than $T_{\rm max}$ were removed.

\bibliography{manuscript_Arxiv}

\end{document}